%% file: 0_main.tex
\pdfoutput=1
\documentclass[11pt]{article}

\usepackage[]{EMNLP2023}

\usepackage{times}
\usepackage{latexsym}

\usepackage[T1]{fontenc}
\usepackage[utf8]{inputenc}

\usepackage{microtype}

\usepackage{inconsolata}

\usepackage{tabularx} %表格自动换行
\usepackage{array}
\usepackage{multirow}
\usepackage{booktabs}
\usepackage{graphicx} 
\usepackage{url} %网页引用
\usepackage{amsthm,amsmath,amssymb} %希腊字母
\usepackage{mathrsfs} %花体字
\usepackage{bm} %公式中的加粗
\usepackage{pifont} %公式需要
\usepackage{float}
\usepackage{lipsum}
\usepackage{soul}
\usepackage{caption}
\usepackage{subcaption}
\usepackage{enumitem}
\usepackage{bbm}
\usepackage{dsfont}
\usepackage{color}
\usepackage{colortbl}
\usepackage{xcolor}
\title{Lion: Adversarial Distillation of Proprietary Large Language Models}

\newcommand*{\affaddr}[1]{#1} % No op here. Customize it for different styles.
\newcommand*{\affmark}[1][*]{\textsuperscript{#1}}
\newcommand*{\email}[1]{\texttt{#1}}
\newcommand*{\equalcontribution}{\textsuperscript{*}}

\author{
Yuxin Jiang\affmark[1,2]
~~~~Chunkit Chan\affmark[2]\equalcontribution
~~~~Mingyang Chen\affmark[1,2]\equalcontribution
~~~~Wei Wang\affmark[1,2]\\
\affaddr{\affmark[1]The Hong Kong University of Science and Technology (Guangzhou), Guangzhou, China}\\
\affaddr{\affmark[2]The Hong Kong University of Science and Technology, Hong Kong SAR, China}\\
\email{\{yjiangcm, ckchancc, mchenbt\}@connect.ust.hk},
\email{weiwcs@ust.hk} \\
}

\begin{document}
\maketitle

% Add the footnote for equal contributions
\begingroup
\renewcommand\thefootnote{$*$}
\footnotetext{The two authors have equal contributions.}
\endgroup

\begin{abstract}
The practice of transferring knowledge from a sophisticated, proprietary large language model (LLM) to a compact, open-source LLM has garnered considerable attention.
Previous works have focused on a unidirectional knowledge distillation way by aligning the responses of the student model with those of the teacher models to a set of instructions.
Nevertheless, they overlooked the possibility of incorporating any ``feedback''\textemdash identifying challenging instructions where the student model's performance falls short\textemdash to boost the student model’s proficiency iteratively.
% from the student model, potentially undermines the efficacy and efficiency of the learning processes.
To this end, we propose a novel adversarial distillation framework for a more efficient knowledge transfer.
Leveraging the versatile role adaptability of LLMs, we prompt the teacher model to identify ``hard'' instructions and generate new ``hard'' instructions for the student model, creating a three-stage adversarial loop of imitation, discrimination, and generation.
By applying this adversarial framework, we successfully transfer knowledge from ChatGPT to a student model (named \textbf{Lion}), using a mere 70k training data.
Our results show that Lion-13B not only achieves comparable open-ended generation capabilities to ChatGPT but surpasses conventional state-of-the-art (SOTA) instruction-tuned models like Vicuna-13B by 55.4\% in challenging zero-shot reasoning benchmarks such as BIG-Bench Hard (BBH) and 16.7\% on AGIEval.\footnote{Code and model can be found at \url{https://github.com/YJiangcm/Lion}.}

% , achieving nearly 95\% capability approximation using a mere 70k training data.
% We aspire that this proposed model may serve as the baseline to reflect the performance of ChatGPT, especially the open-source instruction-following language model baseline for our community.

\end{abstract}

\input{1_introduction.tex}

\input{2_related_work.tex}

\input{3_method.tex}

\input{4_experiments.tex}

\input{5_conclusion.tex}

\input{6_limitations.tex}

\newpage

\bibliography{anthology}
\bibliographystyle{acl_natbib}

\input{7_appendix.tex}

\end{document}

%% file: 1_introduction.tex
\section{Introduction}

% The recent advancements in Natural Language Processing (NLP) literature underscore a remarkable surge in buliding models capable of following natural language instructions.
Large language models (LLMs) capable of following natural language instructions have exhibited tremendous success in generalizing zero-shot to new tasks \cite{DBLP:conf/acl/MishraKBH22, DBLP:conf/iclr/WeiBZGYLDDL22}. Due to various concerns, the most advanced LLMs, such as ChatGPT \cite{openai2022chatgpt} and GPT-4~\cite{DBLP:journals/corr/abs-2303-08774} that boasting billions of parameters, are typically proprietary, comprising both the model parameter and the training data. 
To foster increased transparency regarding their intricate operational mechanics, a surge in research efforts focusing on knowledge distillation from a proprietary ``teacher'' LLM to an open-source ``student'' LLM.
This is typically accomplished by aligning the responses of the student model with those of the teacher model to a set of instructions, which can be manually or automatically generated \cite{wang2022selfinstruct, alpaca, vicuna2023, xu2023wizardlm}.

% The most advanced LLMs boasting billions of parameters, including ChatGPT \cite{openai2022chatgpt} and GPT-4~\cite{DBLP:journals/corr/abs-2303-08774}, are typically closed-source (including both the model parameter and the training data) due to various concerns. 

\renewcommand{\dblfloatpagefraction}{.9}
\begin{figure}[]
\centering
\includegraphics[width=\linewidth]{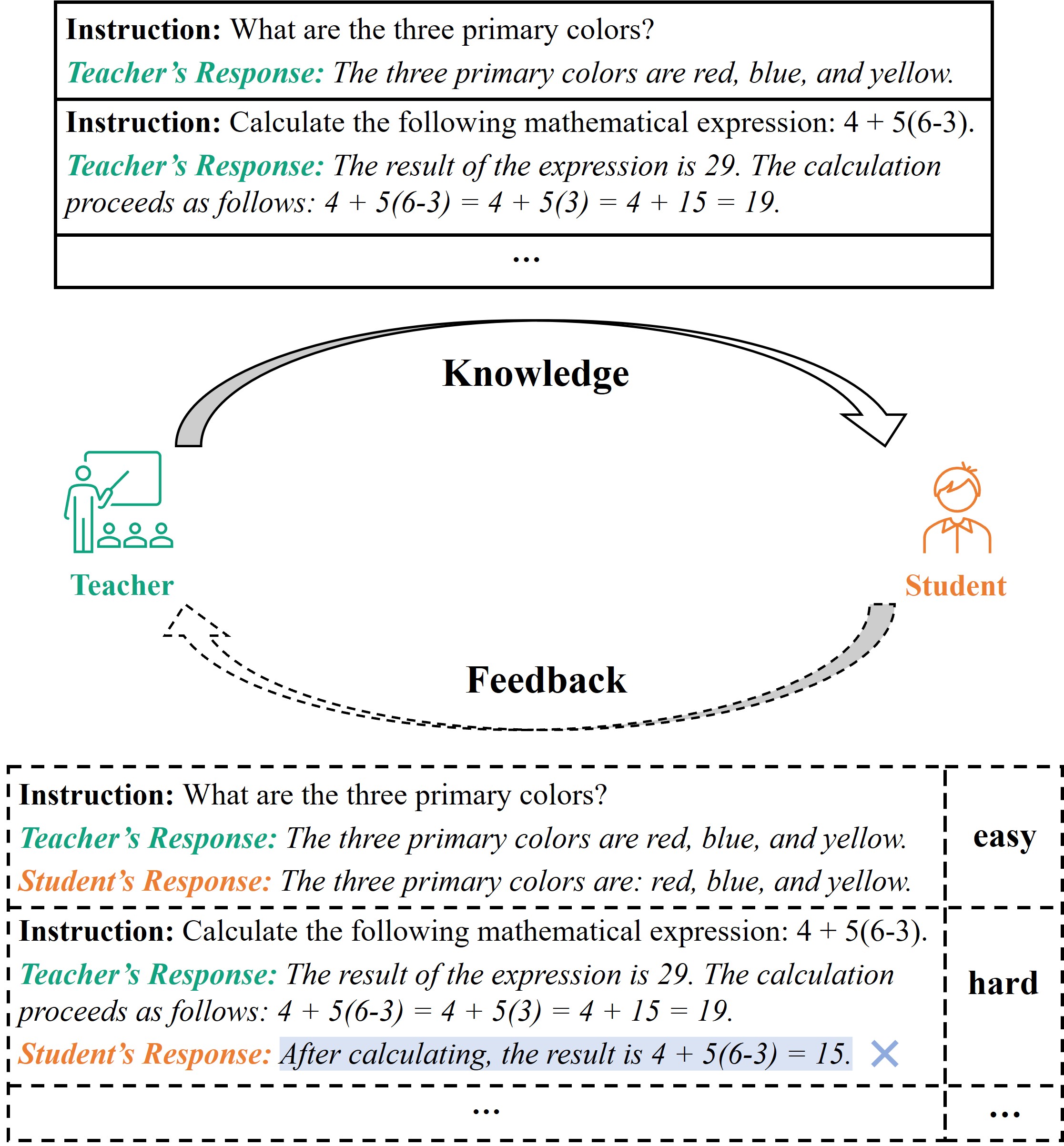}
\caption{An illustration of the distinction between our approach and earlier ones. Previous methods facilitate a one-way knowledge transfer from the teacher to the student (\textit{solid arrow}). Our approach, however, incorporates an innovative step (\textit{dashed arrow}) that completes a loop: it enables the feedback”\textemdash identifying the student model’s weaknesses\textemdash to be relayed back to the teacher, in order to foster tailored learning.}
\label{fig:intro}
\vspace{-0.5 cm}
\end{figure}

However, previous works employ a unidirectional approach to knowledge transfer (solid arrow in Figure \ref{fig:intro}), where the teacher imparts knowledge to the student without considering any ``feedback''.
To better illustrate this using a tangible classroom scenario, the ``feedback'' refers to identifying the ``hard'' examples or problems where the student's performance falls short.
This feedback guarantees that the teacher can provide bespoke training that centers on ``hard'' examples, thereby paving the way for more effective and tailored learning experiences for the student.

Inspired by adversarial knowledge distillation (AKD), which aims to iteratively improve the student model's performance by learning from generated hard samples \cite{fang2019akd, DBLP:conf/nips/MicaelliS19, DBLP:conf/aaai/HeoLY019}, we propose an adversarial framework for distilling a proprietary LLM into a compact student model.
Nevertheless, these AKD methodologies necessitate accessibility to the weights or gradients of the
teacher model, which cannot be directly adapted to our setting.
% to discriminate hard samples
To circumvent this problem, we leverage the unparalleled role adaptability of LLMs, which can be effectively employed through a diverse range of prompts \cite{DBLP:conf/iclr/SanhWRBSACSRDBX22}. 
In particular, we prompt the proprietary teacher LLM to serve as a ``referee'' to discriminate hard instructions where there exists a significant performance discrepancy between the teacher's and student's responses, and serve as a ``generator'' to produce new instructions that emulate the data distributions corresponding to the discriminated hard instructions. 
Our framework, as depicted in Figure \ref{fig:overview}, consists of three stages in an iteration: 1) an imitation stage to align the student's response with the teacher's response; 2) a discrimination stage to identify hard instructions; 3) A generation stage to produce new hard instructions for escalating the challenges presented to the student model.
In essence, our adversarial framework forms a \emph{positive feedback loop} that efficiently bootstraps the student model's proficiency.

To verify the efficiency and efficacy of our method, we apply our AKD framework to transfer the knowledge of ChatGPT \footnote{We access ChatGPT using the OpenAI API (\textit{gpt-3.5-turbo model}).} onto an open-source foundation LLM, known as LLaMA \cite{touvron2023llama}.
We select Alpaca's training data (generated from only 175 manually selected seed instructions) as the initial training instructions and execute three iterations of AKD, resulting in a total of 70K data that our model is trained on.
We've christened our model as \textbf{Lion}, drawing inspiration from the art of ``distillation''.
% Through comprehensive experiments on open-ended generation and reasoning datasets (containing 40 sub-tasks in total), our Lion-13B showcases superior performance against competitive instruction-tuned baselines like Vicuna. 
By conducting extensive experiments on open-ended generation and reasoning datasets, which include a total of 40 sub-tasks, our Lion-13B showcases superior performance surpassing instruction-tuned baseline models such as Vicuna~\cite{vicuna2023}.
% To make a fair comparison with previous baselines, including LLaMA \cite{touvron2023llama}, Alpaca \cite{alpaca}, Vicuna \cite{vicuna2023}, and WizardLM \cite{xu2023wizardlm},
% we conducted an automatic evaluation with GPT-4 and a human evaluation with alignment criteria by following previous works. 
% The experimental results demonstrate that our 7B model, Lion, showcases superior performance across both evaluation paradigms.
% Strikingly, Lion 7B is shown to acquire nearly 95\% capability of ChatGPT, as automatically accessed by GPT-4.
Our main contributions are as follows:
\begin{itemize}[]
\item Our work is the first attempt to adopt the idea of adversarial knowledge distillation to large language models.
\item Our proposed framework demonstrates impressive efficiency and efficacy. With instruction tuning performed on 70k data without any human annotation, our Lion-13B approximates ChatGPT's capabilities on open-ended generation dataset and largely outperforms the current SOTA model Vicuna-13B on reasoning tasks.
\item The versatility of our framework allows for broad application: it is not exclusive to ChatGPT but can be conveniently adapted to suit a variety of other proprietary LLMs.
% \item We release the compact student model, \textbf{Lion}, as an accessible and lightweight springboard for future research endeavors. 
\end{itemize}

% \cite{vicuna2023} fine-tune a student model on 70K user-shared multi-turn conversions (around 700K instruction-following samples) with ChatGPT from ShareGPT.com.

% From the perspective of energy-based learning, we try to make the student model and ChatGPT compatible on the data distribution and push down the examples that have large energy values on the energy surface (i.e., lack of compatibility with ChatGPT).

%https://crfm.stanford.edu/2023/03/13/alpaca.html

% Show the differences and limitations:

% Vicuna \cite{vicuna2023}:
% Vicuna collects 70K user-shared multi-turn conversions (around 700K instruction-following samples) with ChatGPT from ShareGPT.com.

% WizardLM \cite{xu2023wizardlm}:
% 1. Our method is DFS (more efficient), and WizardLM is BFS.
% 2. WizardLM manually designs some rules to generate hard examples; while our method can do this automatically.

%% file: 2_related_work.tex
\section{Related Work}

\subsection{Instruction-Following Language Models}
With the impressive ability of instruction-following large language models such as ChatGPT~\cite{openai2022chatgpt} and GPT-4~\cite{DBLP:journals/corr/abs-2303-08774},
% has been exhibited by many studies~\cite{DBLP:journals/corr/abs-2206-07682, 
% DBLP:journals/corr/abs-2303-12712,
% DBLP:journals/corr/abs-2302-04023, 
% DBLP:journals/corr/abs-2302-10724,
% DBLP:journals/corr/abs-2304-14827}, 
% the techniques such as instruction tuning~\cite{wei2022finetuned} and reinforcement learning from human feedback (RLHF)~\cite{ouyang2022training} have attracted a lot of attention.
the techniques of instruction tuning~\cite{wei2022finetuned} have attracted a lot of attention~\cite{DBLP:journals/corr/abs-2206-07682, 
DBLP:journals/corr/abs-2303-12712,
DBLP:journals/corr/abs-2302-04023, 
DBLP:journals/corr/abs-2302-10724,
DBLP:journals/corr/abs-2304-14827}.
The early research of instruction tuning aims to enhance the generalization ability of language models, allowing these models to perform new tasks by comprehending task descriptions without relying on a few examplars. By fine-tuning these instruction-following language models (e.g., T5~\cite{DBLP:journals/jmlr/RaffelSRLNMZLL20}, FLAN~\cite{DBLP:conf/iclr/AribandiTSRZMZ022}, T0~\cite{DBLP:conf/iclr/SanhWRBSACSRDBX22}, and ExT5~\cite{DBLP:conf/iclr/AribandiTSRZMZ022}) on multi-task datasets in the form of natural language phrased as instructions, these models have been shown to perform well on unseen tasks with the instructions. 
% In addition, ZeroPrompt~\cite{DBLP:conf/emnlp/XuCDSWLY22a} and Flan-T5~\cite{DBLP:journals/corr/abs-2210-11416} expanded the number of tasks to thousands with the designed instruction for each task. The performance of these models consistently improves on the downstream tasks.

However, these models are only fine-tuned on simple task-specific instructions, and it is challenging to comprehend the sophisticated and diverse intent of users in real-world scenarios. Therefore, InstructGPT~\cite{wei2022finetuned}, ChatGPT~\cite{openai2022chatgpt}, and GPT-4~\cite{DBLP:journals/corr/abs-2303-08774} trained on the diverse forms and abundant task types of human-crafted instructions annotated by a considerable number of annotators. Since these instructions were not open-sourced, recent works such as Alpaca~\cite{alpaca}, Vicuna~\cite{vicuna2023}, and WizardLM~\cite{xu2023wizardlm} investigate how to generate high-quality instructions and fine-tune the open-source large language model LLaMA~\cite{touvron2023llama} with them to approach the performance of ChatGPT.

\subsection{Knowledge Distillation}

Knowledge Distillation (KD) \cite{hinton2015distilling,radosavovic2018data,chen2019distilling} represents a crucial strategy within the sphere of model compression and acceleration, wherein a compact student model is instructed to emulate the performance traits of a more cumbersome teacher model. 
In practical contexts, the availability of training data is often constrained due to concerns regarding privacy, legality, security, or confidentiality. To address the absence of training data, data-free KD methods were proposed to align the student model to the teacher model, capitalizing on either related proxy data \cite{orekondy2019knockoff,papernot2017practical} or synthetic data generated by learnable generators (e.g., Generative Adversarial Network (GAN)) \cite{addepalli2020degan,fang2019akd,micaelli2019zero} or teacher model inversions \cite{yin2020dreaming,chawla2021data,fang2022up}. 
Nevertheless, these KD methodologies necessitate the accessibility to the weights or gradients of the teacher model. Consequently, an alternative line of research, commonly denoted as data-free model extraction (or stealing), endeavors to bridge this gap by employing zero-order estimation methodologies to approximate the authentic gradients of the teacher model to guide the update of the optimized generators~\cite{kariyappa2021maze,truong2021data}. 
However, adapting these methods to our distillation task presents two main hurdles. First, these techniques are primarily designed for image-based classification tasks, assuming access to a continuous softmax vector from the teacher model. Estimating zero-order gradients becomes problematic in our case, as responses are typically sequence-oriented. Second, developing an effective instruction generator capable of producing diverse, high-quality instructions that mirror the teacher model's training data distribution proves more challenging than in the image domain.

%% file: 3_method.tex
\section{Methodology}

% 加入COT
% 怎么sampling， continue learning
% mix up. 变成两个问题
% student是白盒，利用起来
% contrastive learning

\renewcommand{\dblfloatpagefraction}{.9}
\begin{figure*}[!tbp]
\centering
\includegraphics[width=0.95\linewidth]{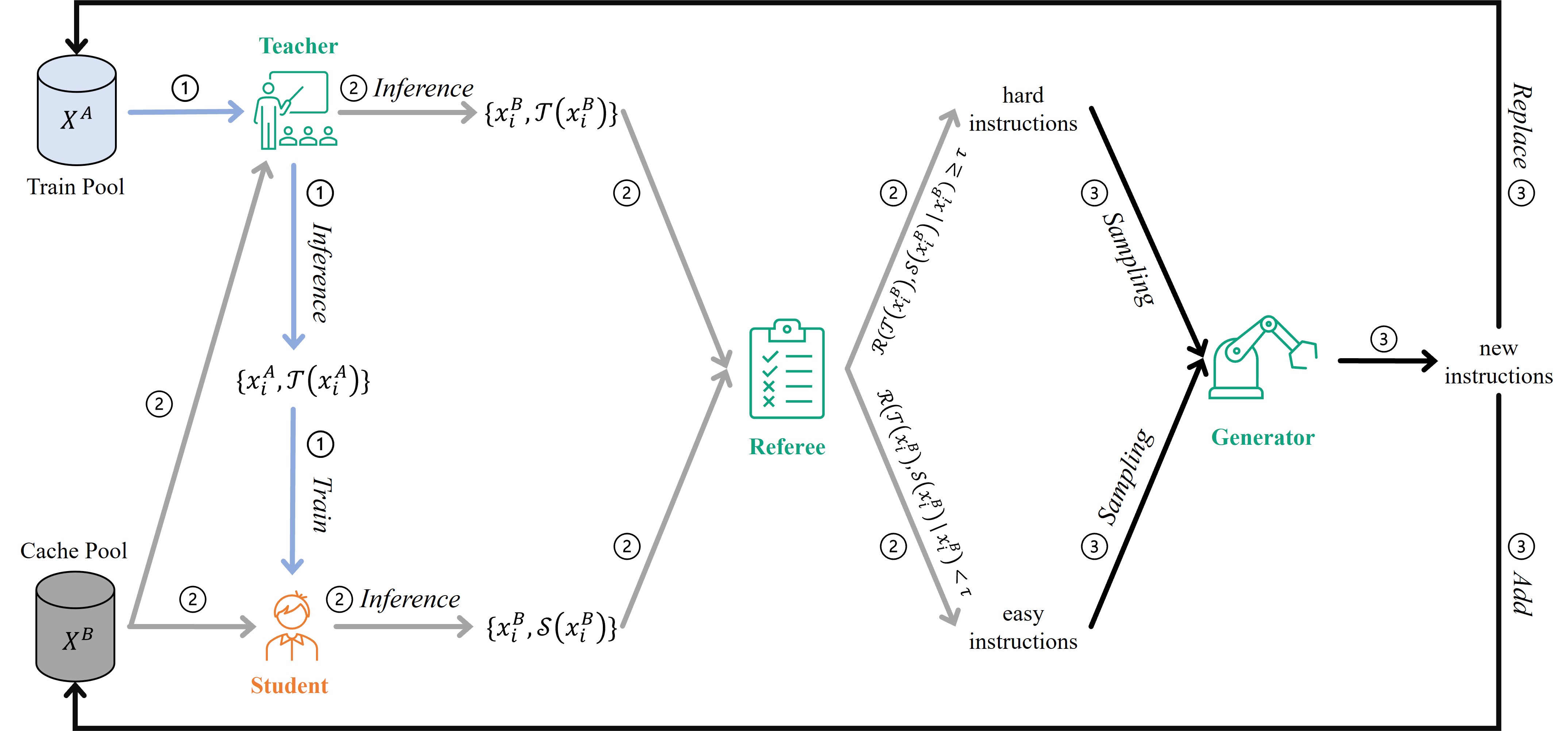}
\caption{The overview of our adversarial distillation framework, where we craft a compact Student LLM $\mathcal{S}$ based on a superior proprietary LLM that serves three roles: the \textbf{Teacher} $\mathcal{T}$, the \textbf{Referee} $\mathcal{R}$, and the \textbf{Generator} $\mathcal{G}$. From left to right, there are three stages in an iteration: 
1) Imitation;
2) Discrimination;
3) Generation.
}
\label{fig:overview}
\end{figure*}

Harnessing the learned knowledge of a sophisticated teacher model $\mathcal{T}(x; \theta^\mathcal{T})$ where the parameter $\theta^\mathcal{T}$ is inaccessible, our goal is to craft a more lightweight student model $\mathcal{S}(x; \theta^\mathcal{S})$.
% without any access to the training data of the teacher model.
Ideally, a student model is optimal if the expectation of model discrepancy (which indicates the prediction differences between teacher $\mathcal{T}$ and student $\mathcal{S}$) on the uniform data distribution is minimized. 
Inspired by the success of adversarial knowledge distillation (AKD) \cite{fang2019akd, DBLP:conf/nips/MicaelliS19, DBLP:conf/aaai/HeoLY019}, we turn to optimize an upper bound of the expectation \textemdash the expectation of the model discrepancy on ``hard samples'', where the teacher $\mathcal{T}$ and the student $\mathcal{S}$ have a relatively large performance gap. 
These ``hard samples'' are inclined to dominate the expectation of the model discrepancy.
Thus, the overall expected model discrepancy can be effectively and efficiently reduced by optimizing the student model $\mathcal{S}$ on these ``hard samples''.
% approximate the model $\mathcal{T}$ with a parameterized $\mathcal{S}$ by iteratively minimizing the model discrepancy on "hard samples", where the teacher $\mathcal{T}$ and the student $\mathcal{S}$ have a relatively large performance gap.
% The intuition behind is quite straightforward by considering a real-world scenario: the most efficient way to improve a student's capability is to make the student focus on the "hard" knowledge.
The underlying rationale is rather straightforward and can be analogized to a real-world educational scenario: continuously concentrating on the ``hard'' knowledge that the student finds challenging to grasp is the most effective manner of enhancing a student's proficiency.

However, in the process of training the student model $\mathcal{S}$, hard samples will be mastered by the student and converted into easy samples. 
Hence we need a mechanism to continuously generate hard samples, which can be achieved by an adversarial framework.

The whole framework of our \textit{Adversarial Knowledge Distillation} is depicted in Figure \ref{fig:overview},
which contains three stages in an iteration: 
% 1) An imitation stage to train Student using Teacher’s knowledge;
% 2) A discrimination stage to identify hard samples;
% 3) A generation stage to produce new hard samples to enlarge the training data.
1) an imitation stage to align the student's response with the teacher's response; 
2) a discrimination stage to identify hard samples; 
3) A generation stage to produce new hard samples for escalating the challenges presented to the student model.

\subsection{Initilization}
As shown in Figure \ref{fig:overview}, four roles and two data pools are established in our framework, and we will comprehensively illustrate their functions later.
We initialize our student model $\mathcal{S}$ using a foundation LLM such as LLaMA \cite{touvron2023llama}. We initialize our teacher model $\mathcal{T}$, referee $\mathcal{R}$, and generator $\mathcal{G}$ by using the same proprietary LLM such as ChatGPT \cite{openai2022chatgpt}.
The multiple roles that this proprietary LLM serves are accomplished through the use of varied prompt templates.
% The Train Pool $X^{A}$ comprises instructions to fine-tune our student model $\mathcal{S}$ using the knowledge from the teacher model $\mathcal{T}$.
% The Cache Pool $X^{B}$ consists of instructions to evaluate the performance of $\mathcal{S}$ and $\mathcal{T}$ for further identification of hard instructions.
We start the iteration from a given initial Train Pool $X^{A}=\{x^{A}_i\}_{i \in [1, N^A]}$, where $x^{A}_i$ is the $i$-th instruction in $X^{A}$, and $N^A$ is the number of samples in $X^{A}$.
The Cache Pool $X^{B}$ is initialized as identical to $X^{A}$, consisting of instructions to evaluate the performance of $\mathcal{S}$ and $\mathcal{T}$.
% Note that the instruction here is composed of an instruction prompt and an instance input, and there does not have a strict boundary between them in many cases.
% For example, ``write an abstract about knowledge distillation'' can be a valid instruction prompt that we expect models to respond to directly, while it can also be formulated as ``write an abstract about the following method'' as the instruction prompt, and ``knowledge distillation'' as an instance input. To encourage the data format's diversity, we allow instructions that do not require additional input.

\subsection{Imitation Stage}
% In the standard practice of knowledge distillation applied to language models, the most prevalent loss function employed is the Kullback-Leibler Divergence, which quantifies the difference between the output probability distributions of the teacher model $\mathcal{T}$ and the student model $\mathcal{S}$.
% However, in our setting, the teacher $\mathcal{T}$ behaves as a black-box model, meaning its output probability distribution is not accessible.
To impart the knowledge of the teacher to the student, we construct the instruction-response data $\{x^{A}_i, \mathcal{T}(x^{A}_i)\}_{i \in [1, N^A]}$ by forward propagating instructions in the Train Pool $X^{A}$ through the teacher $\mathcal{T}$.
The prompt template used for model inference is shown in Table \ref{tab:response}.
% With this training data, where $x^{A}_i$ serves as the input and $\mathcal{T}(x^{A}_i)$ as the target, we fine-tune our student model $\mathcal{S}$ by optimizing the language modeling objective.
Like the imitation training of previous work~\cite{alpaca, vicuna2023}, we fine-tune our student model $\mathcal{S}$ to align the response of the teacher model, by optimizing the autoregressive language modeling objective.

% In the vanilla knowledge distillation of language models, the most commonly used loss function is the Kullback-Leibler Divergence, which quantifies the difference between the output probability distribution of teacher $\mathcal{T}$ and that of student $\mathcal{S}$. 
% Since our $\mathcal{T}$ is a black-box model whose output probability distribution is inaccessible, we forward propagate instructions in the Train Pool $X^{A}$ through $\mathcal{T}$ to acquire the instruction-response data pair $(X^{A}, \mathcal{T}(X^{A}))$.

% Another essential purpose of the imitation stage, in addition to learning knowledge from the teacher, is to construct a better search space for the following stages to find new hard samples.

\begin{figure*}[!tbp]
  \centering
  \begin{subfigure}[b]{0.27\linewidth}
    \includegraphics[width=\linewidth]{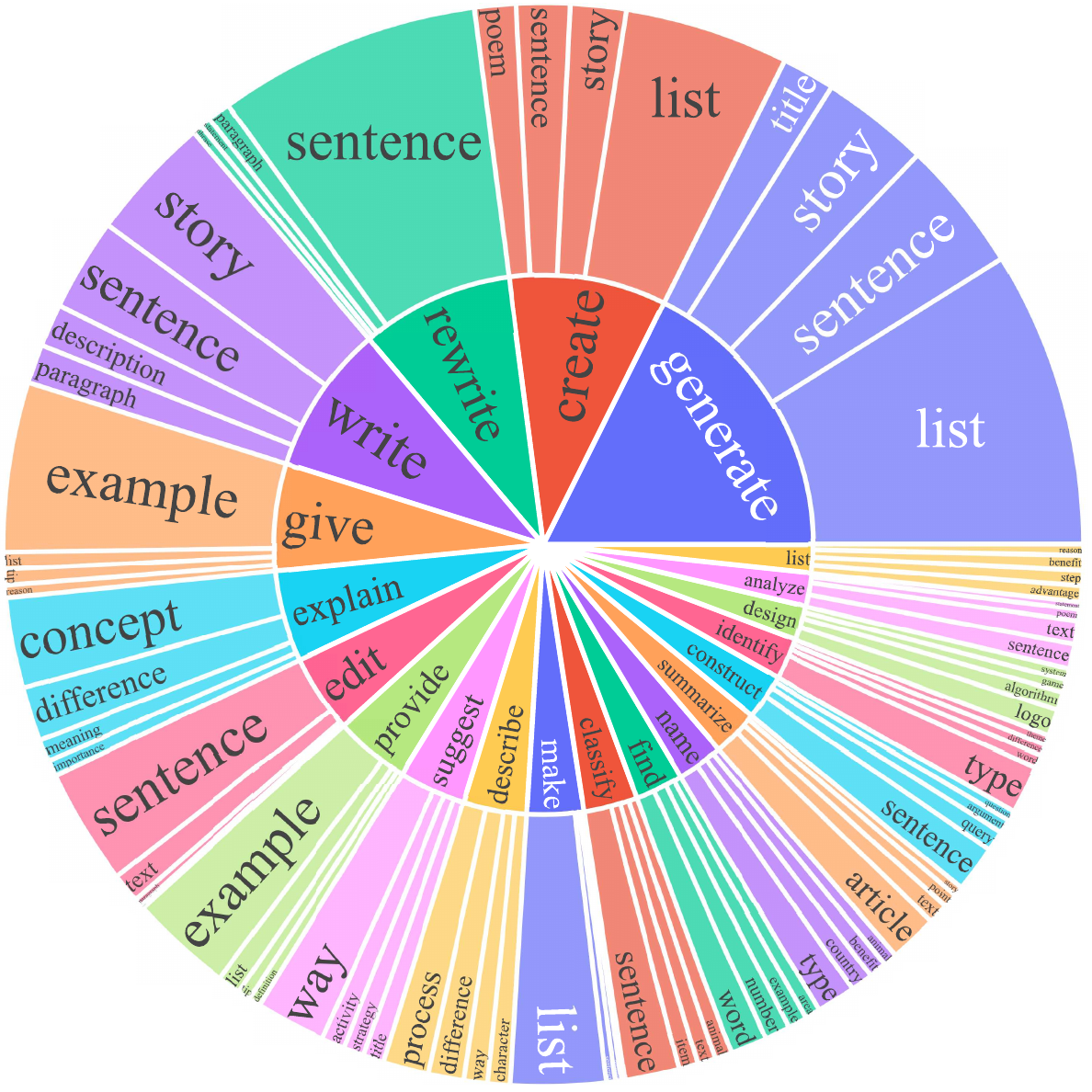}
    \caption{Instructions of the Cache Pool in the first iteration.}
    \label{fig:model1}
  \end{subfigure}
  \qquad % add some horizontal spacing between the subfigures
  \begin{subfigure}[b]{0.27\linewidth}
    \includegraphics[width=\linewidth]{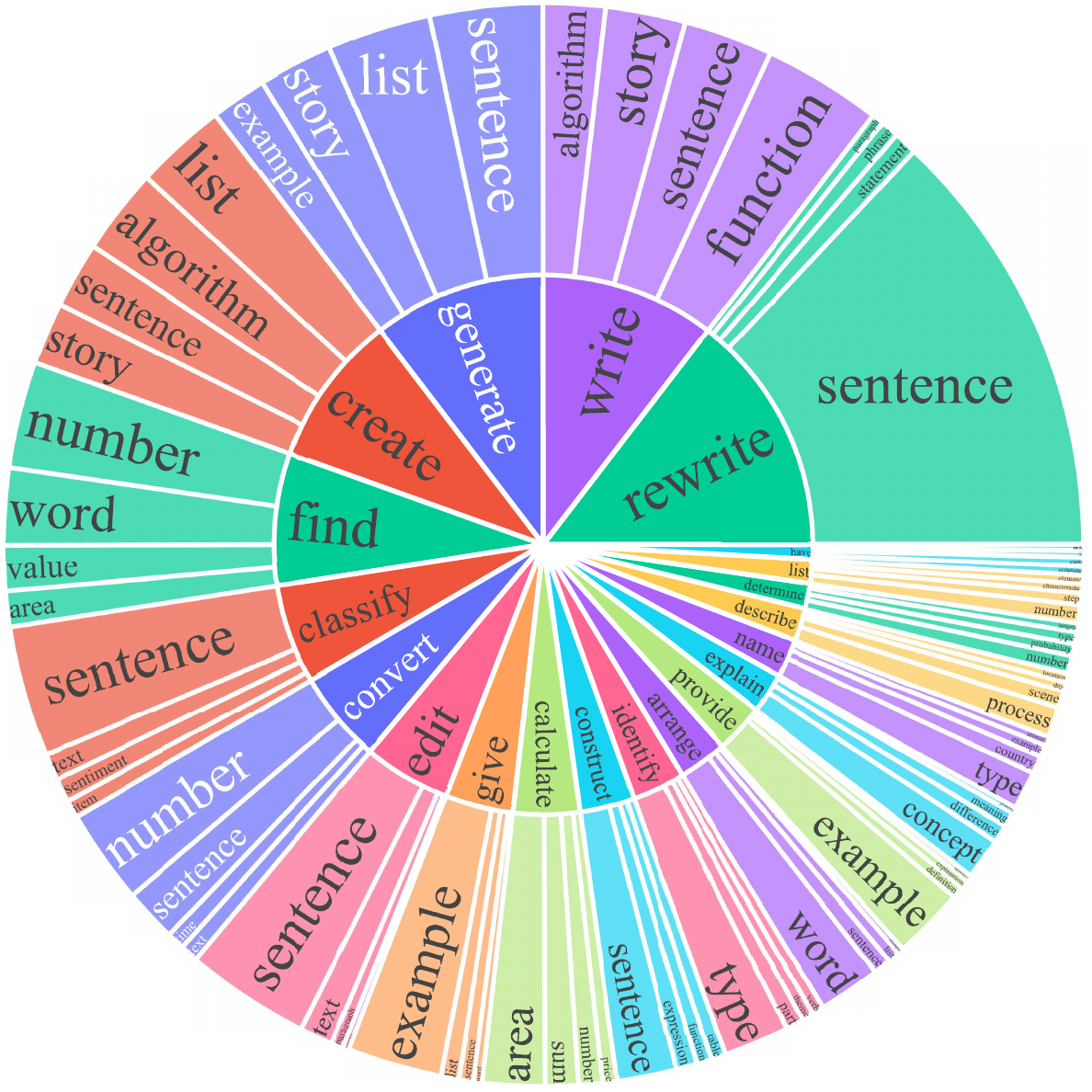}
    \caption{Identified hard instructions in the first iteration.}
    \label{fig:model2}
  \end{subfigure}
  \qquad % add some horizontal spacing between the subfigures
  \begin{subfigure}[b]{0.27\linewidth}
    \includegraphics[width=\linewidth]{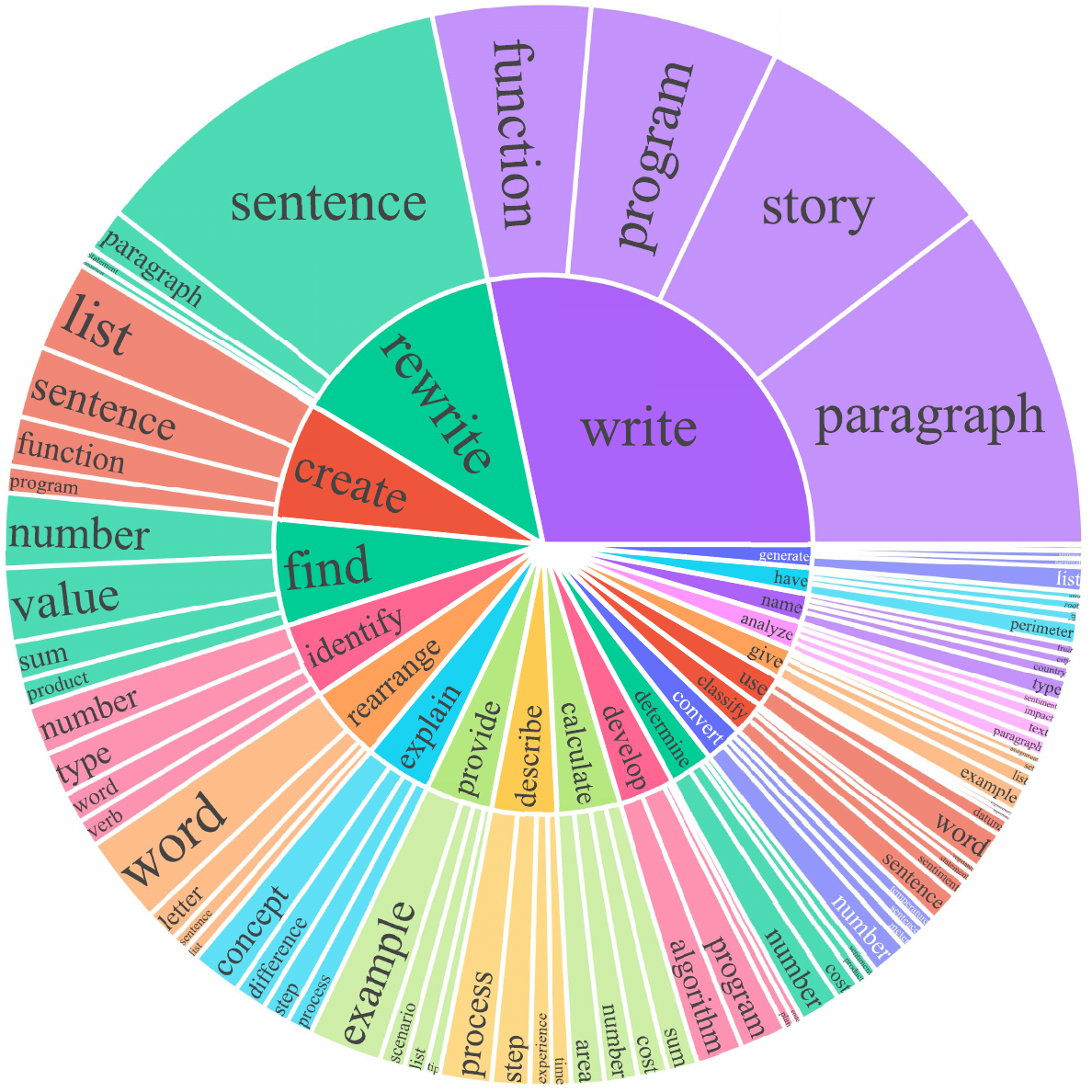}
    \caption{Generated hard instructions in the first iteration.}
    \label{fig:model3}
  \end{subfigure}
  \caption{The top 20 most common root verbs (inner circle) and their top 4 direct noun objects (outer circle) in the instructions.}
  \label{fig:comparison}
  % \vspace{-1 cm}
\end{figure*}

\subsection{Discrimination Stage}
\label{sec:discrimination}
Figure \ref{fig:overview} demonstrates that the discrimination stage starts from the Cache Pool, denoted as $X^{B}$. Even though this pool begins with the same initialization as the Train Pool, their uses diverge. The Train Pool is rejuvenated by replacing its existing instructions with freshly generated instructions, whereas the Cache Pool is enriched by incorporating these generated instructions. As a result, the growing storage capacity of the Cache Pool provides a more extensive space for evaluating the performance gap between teacher $\mathcal{T}$ and student $\mathcal{S}$. This allows for more thorough detection of hard instructions.

% Recent studies have demonstrated that large language models, such as ChatGPT, have the potential to serve as unbiased referees in gauging the quality of responses produced by two different AI assistants \cite{openai2022chatgpt, vicuna2023}. 
% % The evaluation could take several factors into account: semantic accuracy, contextual relevance, grammatical correctness, and fluency. 
% Language models are trained on vast corpora of text and thus have a broad understanding of language use, style, and grammar, making them well-suited to evaluate the linguistic quality of the responses. In terms of content, the breadth of a language model's training data enables it to judge the factual correctness and contextual relevance of the responses.
In the discrimination stage, we ask the proprietary LLM to serve as a ``referee'', which quantifies the performance gap between $\mathcal{T}$ and $\mathcal{S}$.
Specifically, we feed each instruction $x^{B}_i$ in the Cache Pool $X^{B}$ through both the teacher $\mathcal{T}$ and student $\mathcal{S}$ to generate the outputs $\mathcal{T}(x^{B}_i)$ and $\mathcal{S}(x^{B}_i)$, respectively.
Then we ask the referee $\mathcal{R}$ to quantitatively measure the quality difference between teacher's response $\mathcal{T}(x^{B}_i)$ and student's response $\mathcal{S}(x^{B}_i)$, conditioned on $x^{B}_i$:
\begin{equation}
d_i = \mathcal{R}(\mathcal{T}(x^{B}_i), \mathcal{S}(x^{B}_i) \ | \ x^{B}_i) 
\end{equation}
The above process is conducted by using the prompt template (as shown in Table \ref{tab:discriminator_gpt-3.5-turbo}) inspired by \cite{vicuna2023}, which requires the LLM to consider the helpfulness, relevance, accuracy, and level of detail of two responses and output two scores. 
To mitigate the positional bias~\cite{DBLP:journals/corr/abs-2305-17926} of the LLM referee, we conduct two runs by exchanging the positions of the teacher's response and the student's response and compute the final score as the average of the two runs.
Then $d_i$ is calculated as the difference between the teacher's score and the student's score.
By setting a threshold $\tau$ (1.0 used in our experiments), we discriminate hard instructions as those instructions with $d_i \ge \tau$, and the others are identified as easy ones.
Figure \ref{fig:model2} provides a clear and intuitive demonstration of which kinds of instructions are discriminated as hard in the first iteration.
Compared with the instructions in the Cache Pool (Figure \ref{fig:model1}), the distribution of the identified hard instructions is quite different, focusing more on complex tasks such as math, coding, etc.

\subsection{Generation Stage}
After carefully discerning the hard instructions, the generation stage aims to produce samples that mirror the data distributions corresponding to these challenging directives.
This process is achieved by employing the proprietary LLM as a generator, denoted as $\mathcal{G}$, leveraging its exceptional prowess in content creation.
Inspired by \cite{xu2023wizardlm}, we randomly sample an instruction from the hard instructions and prompt the generator $\mathcal{G}$ to generate a new instruction.
The newly generated instruction is required to pertain to the same domain and match the task type of the sampled instruction. The template utilized for this prompt is exhibited in Table \ref{tab:generator_hard}.
As shown in Figure \ref{fig:model3}, the distribution of the newly generated hard instructions appears to be comparable to that of the previously identified hard instructions.
To mitigate the issue of catastrophic forgetting and to augment the diversity of the generated instructions, we also randomly sample an instruction from the easy instructions and prompt the generator $\mathcal{G}$ to generate a new instruction that belongs to the same domain as the sampled one, but exhibit a more long-tailed distribution.
The template we use to prompt this process is displayed in Table \ref{tab:generator_no_hard}.
% The examples that demonstrate the generation of both hard and easy instructions are depicted visually in Figure \ref{fig:generation}.

In each iteration, we define $N$ as the total count of newly generated instructions and maintain a 1:1 ratio $r$ between the generated hard instructions and the generated easy instructions.
To promote diversity, a new instruction will be deemed valid only if its ROUGE-L overlap with any existing instructions in the Cache Pool is below 0.7.
Finally, as aforementioned in Section \ref{sec:discrimination}, we proceed to rejuvenate the Train Pool, replacing its existing instructions with freshly generated ones. Concurrently, we enrich the Cache Pool by incorporating these newly generated instructions.

% \begin{figure}[ht]
%     \centering
%     \begin{subfigure}[b]{0.45\textwidth}
%         \includegraphics[width=\textwidth]{emnlp2023-latex/figures/generate_hard.jpg}
%         \caption{Generation of new hard instruction.}
%         \label{fig:sub1}
%     \end{subfigure}
%     \par\bigskip % 或者你可以使用 \newline 命令
%     \begin{subfigure}[b]{0.45\textwidth}
%         \includegraphics[width=\textwidth]{emnlp2023-latex/figures/generate_easy.jpg}
%         \caption{Generation of new easy instruction.}
%         \label{fig:sub2}
%     \end{subfigure}
%     \caption{Output examples of the generator $\mathcal{G}$ given input examples.}
%     \label{fig:generation}
% \end{figure}

\subsection{Min-Max Game Interpretation}
Our adversarial knowledge distillation framework can be interpreted as a dynamic min-max game: in the imitation stage, we fine-tune our student to \textit{minimize} the model discrepancy between itself and the teacher on hard samples; in the discrimination and generation stage, we craft new hard samples to \textit{maximize} the model discrepancy, based on the learning progress of the student model.
This dialectic framework propels the student model towards uncovering otherwise hidden knowledge, paving the way to complete understanding.
As the training progresses through several iterations, the system should ideally achieve equilibrium. This is the point where the student model has mastered all the hard samples and the referee $\mathcal{R}$ can no longer distinguish between the student $\mathcal{S}$ and teacher $\mathcal{T}$ models. At this juncture, $\mathcal{S}$ becomes functionally indistinguishable from $\mathcal{T}$.

%% file: 4_experiments.tex
\section{Experiments Setting}

\subsection{Datasets}
% Evaluating the efficacy of LLM across various tasks presents a considerable challenge since different tasks require quite different expertise \cite{wang2022selfinstruct}.
In our experiments, we implemented a comprehensive LLM evaluation protocol that considers a diverse range of abilities, such as writing, coding, commonsense, math, and logical reasoning.
The datasets we utilized can be classified into two main categories: open-ended generation and reasoning.
% This study aims to gauge the performance of LLMs when faced with unseen instructions to scrutinize their aptitude in handling and executing arbitrary tasks. To this end, two prominent datasets have been employed in our experiments:

\subsubsection{Open-ended Generation Datasets}

\paragraph{Vicuna-Instructions} \cite{vicuna2023} 
% is a dataset synthesized by GPT-4 with 80 challenging questions that baseline models find challenging. 
is a set of 80 questions spanning 9 distinct task categories.
% , which has been widely utilized for evaluating the capability of LLMs.
This dataset has gained extensive usage in evaluating the capabilities of LLMs. Within our work, we examine LLMs' performance on this dataset in two different settings:
% The dataset spans nine distinct categories, namely generic, knowledge, roleplay, common-sense, fermi, counterfactual, coding, math, and writing.
% \paragraph{User-Oriented-Instructions} \cite{wang2022selfinstruct} is a manually assembled dataset including 252 instructions, motivated by 71 user-oriented applications such as Grammarly, StackOverflow, and Overleaf. 

\begin{itemize}
    \item \textbf{Setting1:} Following Vicuna~\cite{vicuna2023}, we leverage GPT-4 to automatically assess the quality of responses (rated on a scale of 1 to 10) between a reference model (ChatGPT) and a candidate model. Subsequently, we calculate the candidate model's performance as the percentage of the total score it achieves compared to the reference model.
    \item \textbf{Setting2:} A recent work~\cite{DBLP:journals/corr/abs-2305-17926} pointed out that a systematic bias may exist in the above-mentioned GPT-4 automatic evaluation. To mitigate this, they propose two strategies, namely Multiple Evidence Calibration and Balanced Position Calibration, to obtain closer alignment with human judgments.
\end{itemize}

\subsubsection{Reasoning Datasets}
\paragraph{AGIEval} \cite{zhong2023agieval} is a well-known benchmark that quantifies the reasoning capability of foundation models in the context of human-centric standardized exams, including college entrance exams, math competitions, lawyer qualification tests, etc.
We choose all English multiple-choice questions (8 tasks, 2,546 samples) among AGIEval for our experiments.
The data statistics are shown in Table \ref{tab:agieval_sat}.
\paragraph{BIG-Bench Hard (BBH)} \cite{suzgun2022bbh} consists of a suite of challenging tasks from BIG-Bench \cite{srivastava2022beyond}, designed to assess the capabilities and limitations of large language models.
These are the tasks on which prior language models underperform the average human rater.
We choose all tasks that can be formatted into multiple-choice questions (23 tasks, 5,511 samples) among BBH for our experiments.
The data statistics are shown in Table \ref{tab:bbheval_sat}.

\paragraph{Setting} We evaluate reasoning capabilities under a zero-shot setting without any exemplars and without Chain-of-Thought (CoT).
For both AGIEval and BBH, we use the prompt format and parsing following~\cite{zhong2023agieval, 2023orca}.
Given the free-form response from the generative models, only the first capital character in the response is considered to compare with the gold answer (exact match).
The result we report is accuracy (\%).

\subsection{Baselines}
We select five superior LLMs as baselines, including LLaMA \cite{touvron2023llama}, Alpaca \cite{alpaca}, WizardLM \cite{xu2023wizardlm}, Vicuna \cite{vicuna2023}, and ChatGPT \cite{openai2022chatgpt}.
% It is worth noticing that Vicuna has been the leading open-source language model in multiple leaderboards including Chatbot Arena. Thus, we make a major comparison with Vicuna.
It is worth noting that Vicuna has consistently ranked as the top open-source language model on multiple leaderboards, such as Chatbot Arena\footnote{\url{https://chat.lmsys.org/?arena}}. Therefore, we will conduct a comprehensive comparison with Vicuna.
See detailed descriptions of these baselines in Appendix \ref{appendix:baseline}.

\subsection{Implementation Details}

\paragraph{Training Details}
Our student model is initialized using the pre-trained LLaMA. 
The Train Pool and Cache Pool are initialized with the 52K automatically generated instructions from Alpaca \cite{alpaca}. 
The total number of iterations is set to 3, with 6K newly generated instructions added at each iteration. 
This results in a total of 70K data that our model is trained on in order to make a fair comparison with current SOTA baselines, including WizardLM and Vicuna.
The training hyperparameters are listed in Appendix \ref{sec:gpt_parameter}.

\paragraph{Inference Details}
To draw inferences from Lion and ChatGPT, we calibrated the temperature to 0.7 and set the maximum generation length at 1024. All other parameters adhere to their default settings.
For LLaMA, Alpaca, WizardLM, and Vicuna, we configured their inference parameters in line with the specifications given in their respective original papers.
When engaging with the gpt-3.5-turbo API for various roles, we employ an array of hyper-parameters, the specifics of which can be located in Appendix \ref{sec:gpt_parameter}.

% \begin{figure}[!tbp]
% \centering
% \includegraphics[width=\linewidth]{emnlp2023-latex/figures/intro_comparision.jpg}
% \caption{Relative response quality against ChatGPT (assessed by GPT-4).}
% \label{fig:intro_comparison}
% \end{figure}

\begin{table}[]
\small
\centering
\begin{tabular}{l|c|c|c}
\toprule
\textbf{Model} & \textbf{Setting1} & \textbf{Setting2} & \textbf{Avg.} \\ \midrule
LLaMA-7B       & 58.46             & 59.12             & 58.79        \\
Alpaca-7B      & 69.29             & 67.20             & 68.25        \\
WizardLM-7B    & 89.29             & 86.67             & 87.98        \\
Vicuna-7B      & 87.79             & 89.96             & 88.88        \\
Lion-7B        & \textbf{94.74}             & \textbf{92.88}             & \textbf{93.81}        \\ \midrule
LLaMA-13B      & 69.23             & 68.21             & 68.72        \\
Alpaca-13B     & 76.87             & 74.69             & 75.78        \\
Vicuna-13B     & 92.25             & 92.97             & 92.61        \\
Lion-13B       & \textbf{96.57}             & \textbf{100.18}            & \textbf{98.38}        \\ \bottomrule
\end{tabular}
\caption{Relative response quality (\%) against ChatGPT (assessed by GPT-4) on Vicuna-Instructions.}
\label{tab:gpt4_eval}
\end{table}

\renewcommand{\dblfloatpagefraction}{.9}
\begin{figure}[!tbp]
\centering
\includegraphics[width=\linewidth]{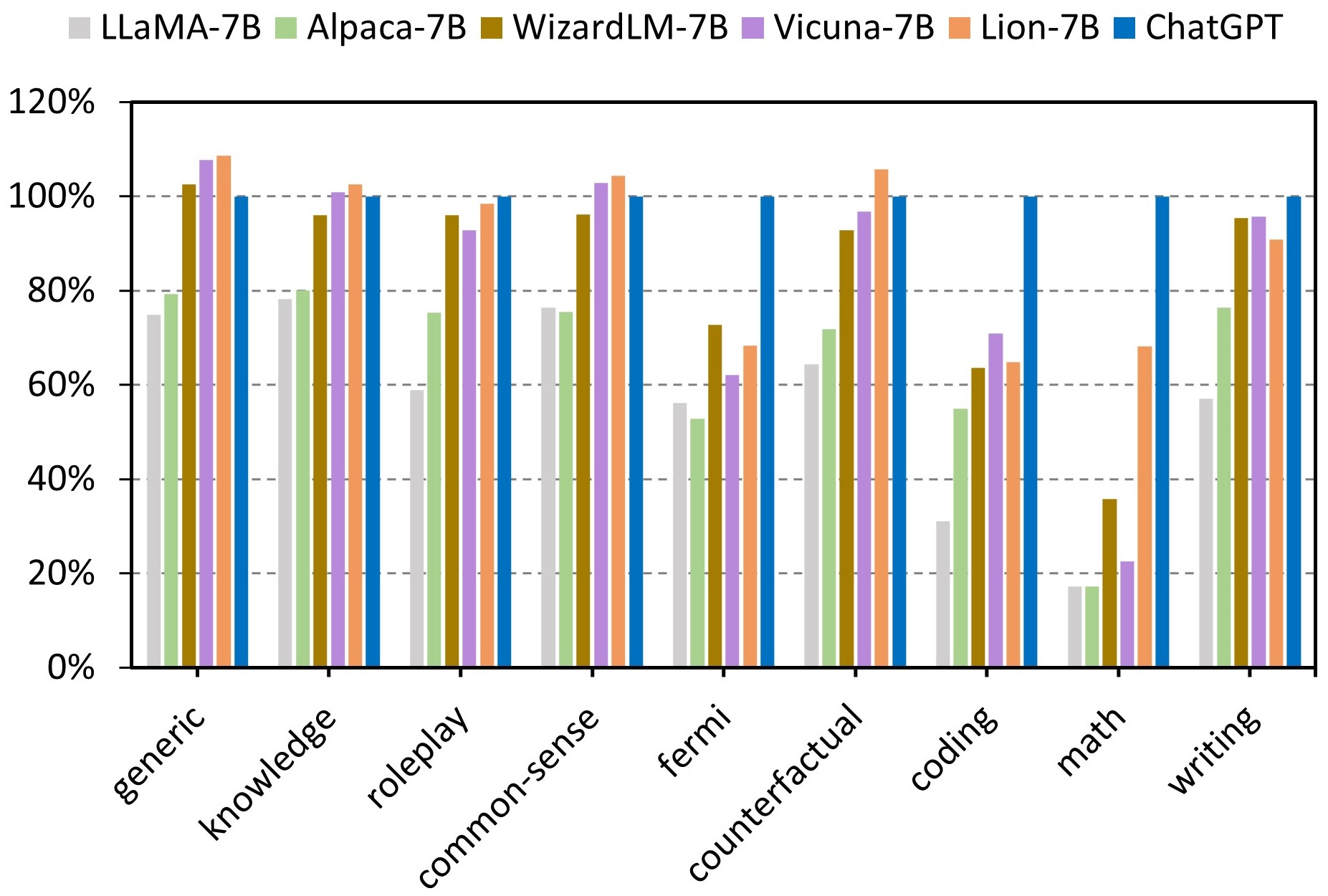}
\caption{
Relative response quality against ChatGPT on diverse task categories of Vicuna-Instructions.
}
\label{fig:80task_category}
% \vspace{-1 cm}
\end{figure}

% \begin{figure}[!tbp]
% \centering
% \includegraphics[width=\linewidth]{emnlp2023-latex/figures/80task_win.jpg}
% \caption{
% The frequency of won, tie, and lost between Lion and baselines on 80 Vicuna-Instructions.
% }
% \label{fig:80task_win}
% \end{figure}

\section{Experimental Results}

\subsection{Results for Open-ended Generation}

Table \ref{tab:gpt4_eval} shows the performance comparison of various models against ChatGPT as the reference model, where GPT-4 is used as a referee/rater.
Our Lion-7B and Lion-13B remarkably outperform their counterparts under two evaluation settings.
Noticeably, Lion-13B shows an 8-point improvement over Vicuna-13B on aggregate, achieving 98.38\% capabilities of ChatGPT.

% \paragraph{Response Quality of Diverse Task Categories}
To comprehensively compare with other baseline models on the capability to generate high-quality responses on various types of instruction, the relative response quality (Setting2) among different task categories is depicted in Figure \ref{fig:80task_category}. Our model impressively and slightly surpasses ChatGPT in the generic, knowledge, common-sense, and counterfactual task categories. Furthermore, for the two difficulty task categories described in the previous study~\cite{vicuna2023, xu2023wizardlm}, our model significantly outperforms other baseline models with at least 32.32\% relative score in the math task category while exceeding most of the baseline in the coding generation task category.

\subsection{Results for Reasoning}

\paragraph{AGIEval Results} Table \ref{tab:agieval} presents the standard zero-shot performance comparison between Lion and baseline models on the AGIEval benchmark for multiple-choice English questions. 
Lion demonstrates significantly stronger performance compared to Vicuna, surpassing it in most task categories and achieving an average relative improvement of over 16\%.
However, Lion-13B still significantly lags behind ChatGPT, only retaining 72.5\% of its reasoning capability.

\paragraph{BIG-Bench Hard Results} Table \ref{tab:bbheval} displays the zero-shot performance comparison between Lion and baseline models on BIG-Bench Hard with standard zero-shot prompting.
Similar to AGIEval, Vicuna exhibits poor performance
on sophisticated reasoning tasks within this benchmark, while Lion substantially surpasses Vicuna by around 50\% on average. 
Particularly, Lion demonstrates significant performance enhancements of over 100\% on tasks involving
data understanding, semantic understanding (Disambiguation QA and Snarks), logical and geometric reasoning (Logical Deduction and Geometric Shapes), and position reasoning (Tracking Shuffled Objects).
Despite achieving an average ability of nearly 74\% compared to ChatGPT on BBH, Lion-13B surpasses ChatGPT in several tasks, including Movie Recommendation, Snarks (identifying sarcastic sentences from two nearly-identical ones), and Tracking Shuffled Objects. This demonstrates the effectiveness of our method.

\renewcommand{\dblfloatpagefraction}{1}
\begin{table*}[]
\footnotesize
\centering
\begin{tabular}{l|ll|l|ll|ll}
\toprule
\multirow{2}{*}{\textbf{Task}} & \multicolumn{2}{c|}{\textbf{Human}} & \multirow{2}{*}{\textbf{ChatGPT}} & \multirow{2}{*}{\textbf{Vicuna-7B}} & \multirow{2}{*}{\textbf{Lion-7B}} & \multirow{2}{*}{\textbf{Vicuna-13B}} & \multirow{2}{*}{\textbf{Lion-13B}} \\
                               & \textbf{Avg}             & \textbf{Top}              &                                  &                                   &                                                                       &               &              \\ \midrule
AQuA-RAT                       & 85.0            & 100.0                                       & 31.9                              & \textbf{23.2}                                                             & 18.5\enspace (-20.3\%)  & 20.1          & \textbf{26.0} \enspace (29.4\%)  \\
LogiQA                         & 86.0            & 95.0                                        & 35.0                              & 21.4                                                             & \textbf{31.8}\enspace (48.6\%) & 29.8  & \textbf{31.3} \enspace (5.0\%)           \\
LSAT-AR                        & 56.0            & 91.0                                     & 24.4                              & \textbf{22.2}                                                            & 17.4\enspace (-21.6\%) & 20.4            & \textbf{23.0} \enspace (12.7\%)  \\
LSAT-LR                        & 56.0            & 91.0                                       & 52.6                              & 18.6                                                            & \textbf{28.2}\enspace (51.6\%)   & \textbf{32.6}          & \textbf{32.6} \enspace (0.0\%)  \\
LSAT-RC                        & 56.0            & 91.0                                        & 65.4                              & 21.9                                                            & \textbf{29.4}\enspace (34.2\%)  & 32.7           & \textbf{40.9} \enspace (25.1\%)  \\
SAT-Math                       & 66.0            & 94.0                                      & 42.7                              & \textbf{21.4}                                                  & 20.9\enspace (-2.3\%)  & 28.6            & \textbf{29.4} \enspace (2.8\%)           \\
SAT-English                    & 66.0            & 94.0                                    & 81.1                              & 25.7                                                           & \textbf{36.4}\enspace (41.6\%)   & 44.2           & \textbf{53.9} \enspace (21.9\%)  \\
SAT-English (w/o Psg.)         & 66.0            & 94.0                                      & 44.2                              & 26.2                                                           & \textbf{27.7}\enspace (5.7\%)  & 26.2            & \textbf{36.2} \enspace (38.2\%)  \\ \midrule
Average                        & 67.1            & 93.8                                       & 47.2                              & 22.6                        & \textbf{26.3}\enspace (16.4\%)      & 29.3                                           & \textbf{34.2} \enspace (16.7\%) \\ \bottomrule
\end{tabular}
\caption{Zero-shot performance comparison of ChatGPT, Vicuna, and Lion on AGIEval (multiple-choice English questions). We report the performance of Human, ChatGPT, and Vicuna from \cite{2023orca}. Performance improvements obtained by Lion over Vicuna are shown in parenthesis.}
\label{tab:agieval}
\end{table*}

\renewcommand{\dblfloatpagefraction}{1}
\begin{table*}[!t]
\footnotesize
\centering
\begin{tabular}{l|l|ll|ll}
\toprule
\textbf{Task} & \textbf{ChatGPT} & \textbf{Vicuna-7B} & \textbf{Lion-7B} & \textbf{Vicuna-13B} & \textbf{Lion-13B} \\ \midrule
Boolean Expressions                   & 82.8             & 39.2                          & \textbf{55.2} \enspace (40.8\%)  & 40.8              & \textbf{65.6} \enspace (60.8\%)     \\
Causal Judgement                      & 57.2             & 39.7                        & \textbf{50.3} \enspace (26.7\%)   & 42.2  & \textbf{43.9} \enspace (4.0\%)             \\
Date Understanding                    & 42.8             & 8.6                    & \textbf{34.0} \enspace (295.3\%)     & 10.0               & \textbf{40.4} \enspace (304.0\%)    \\
Disambiguation QA                     & 57.2             & 15.2                           & \textbf{35.6} \enspace (134.2\%)  & 18.4           & \textbf{44.8} \enspace (143.5\%)    \\
Formal Fallacies                      & 53.6             & 40.0                         & \textbf{46.0} \enspace (15.0\%)  & 47.2             & \textbf{52.4} \enspace (11.0\%)    \\
Geometric Shapes                      & 25.6             & 3.6                         & \textbf{8.8} \enspace (144.4\%)   & 3.6              & \textbf{8.8} \enspace (144.4\%)     \\
Hyperbaton                            & 69.2             & 42.8                        & \textbf{51.6} \enspace (20.6\%)    & 44.0            & \textbf{56.8} \enspace (29.1\%)    \\
Logical Deduction (5 objects)         & 38.8             & 4.8                         & \textbf{19.6} \enspace (308.3\%)  & 4.8              & \textbf{20.8} \enspace (333.3\%)    \\
Logical Deduction (7 objects)         & 39.6             & 1.2                           & \textbf{14.4} \enspace (1100.0\%)   & 1.2           & \textbf{21.2} \enspace (1666.7\%)    \\
Logical Deduction (3 objects)         & 60.4             & 19.6                        & \textbf{40.4} \enspace (106.1\%) & 16.8      & \textbf{38.0} \enspace (126.2\%)             \\
Movie Recommendation                  & 55.4             & 24.4                          & \textbf{26.8} \enspace (9.8\%)    & 43.4          & \textbf{57.6} \enspace (32.7\%)    \\
Navigate                              & 55.6             & 43.6                          & \textbf{49.2} \enspace (12.8\%)  & \textbf{46.4}   & 45.2 \enspace (-2.6\%)             \\
Penguins in a Table                   & 45.9             & 17.5                        & \textbf{24.7} \enspace (41.1\%)    & 15.1            & \textbf{26.7} \enspace (76.8\%)    \\
Reasoning about Colored Objects       & 47.6             & 14.0                        & \textbf{15.2} \enspace (8.6\%)    & 12.0            & \textbf{17.6} \enspace (46.7\%)    \\
Ruin Names                            & 56.0             & 12.2                        & \textbf{14.4} \enspace (18.0\%)    & 15.7            & \textbf{29.2} \enspace (86.0\%)    \\
Salient Translation Error Detection   & 40.8             & 2.0                         & \textbf{12.0} \enspace (500.0\%)   & 2.0             & \textbf{12.4} \enspace (520.0\%)    \\
Snarks                                & 59.0             & 28.0                          & \textbf{56.2} \enspace (100.7\%)   & 28.1           & \textbf{61.2} \enspace (117.8\%)    \\
Sports Understanding                  & 79.6             & 40.4                       & \textbf{48.4} \enspace (19.8\%)    & 48.4             & \textbf{51.6} \enspace (6.6\%)    \\
Temporal Sequences                    & 35.6             & 21.2                         & \textbf{24.4} \enspace (15.1\%)   & \textbf{16.0}   & 10.4 \enspace (-35.0\%)             \\
Tracking Shuffled Objects (5 objects) & 18.4             & 6.4                          & \textbf{14.4} \enspace (125.0\%)    & 9.2           & \textbf{24.8} \enspace (169.6\%)    \\
Tracking Shuffled Objects (7 objects) & 15.2             & 4.0                          & \textbf{13.6} \enspace (240.0\%)  & 5.6    & \textbf{13.2} \enspace (135.7\%)             \\
Tracking Shuffled Objects (3 objects) & 31.6             & 26.8                        & \textbf{34.0} \enspace (26.9\%)    & 23.2            & \textbf{34.4} \enspace (48.3\%)    \\
Web of Lies                           & 56.0             & \textbf{49.4}                       & 47.2 \enspace (-4.5\%)   & 41.2              & \textbf{54.8} \enspace (33.0\%)    \\ \midrule
Average                               & 48.9             & 21.9                         & \textbf{32.0} \enspace (45.9\%)    & 23.3           & \textbf{36.2} \enspace (55.4\%)   \\ \bottomrule
\end{tabular}
\caption{Zero-shot performance comparison of ChatGPT, Vicuna, and Lion on BIGBench Hard (multiple-choice questions) without CoT. We report the performance of ChatGPT and Vicuna from \cite{2023orca}. Performance improvements obtained by Lion over Vicuna are shown in parenthesis.}
\label{tab:bbheval}
\end{table*}

\section{Analyses}

\subsection{Ablation Studies}

\paragraph{The threshold $\tau$ for distinguishing between hard and easy instructions}
We systematically explored $\tau$ ranging from 0.0 to 2.0 and documented its influence on average performance across three datasets. Table \ref{tab:ablation_tau} reveals an optimal range of $\tau$ between 1.0 and 1.5 for all datasets. Notably, elevating $\tau$ from 0.0 to 1.0 consistently enhances performance across all datasets, indicating effective differentiation between hard and easy instructions. However, a continuous increase from 1.0 to 2.0 gradually degrades performance due to decreased diversity in hard instructions. The ablation results demonstrate that our method is not quite sensitive to a large value of $\tau$.

\begin{table*}[]
\small
\centering
\begin{tabular}{c|c|c|c}
\toprule
\textbf{Threshold $\tau$} & \textbf{Vicuna-Instructions (Avg.)} & \textbf{AGIEval (Avg.)} & \textbf{BBH (Avg.)} \\ \midrule
0.0 & 89.58	& 22.4 &	26.5        \\
0.5	& 92.16 &	23.5 &	29.8        \\
1.0	& 93.81	& \textbf{26.3}	& \textbf{32.0}        \\
1.5	& \textbf{94.09}	& 25.7 &	31.6        \\
2.0	& 92.23	& 24.6	& 31.3        \\ \bottomrule
\end{tabular}
\caption{Ablation study of the threshold $\tau$ for Lion-7B.}
\label{tab:ablation_tau}
\end{table*}

\begin{table*}[]
\small
\centering
\begin{tabular}{c|c|c|c}
\toprule
\textbf{Ratio $r$} & \textbf{Vicuna-Instructions (Avg.)} & \textbf{AGIEval (Avg.)} & \textbf{BBH (Avg.)} \\ \midrule
1:0	& 89.60	& 24.3	& 30.8        \\
2:1	& 92.95	& 25.7	& \textbf{33.1}        \\
1:1	& \textbf{93.81}	& \textbf{26.3}	& 32.0        \\
1:2	& 91.77	& 23.9	& 29.6        \\
0:1	& 90.02	& 22.1	& 24.3        \\ \bottomrule
\end{tabular}
\caption{Ablation study of the ratio $r$ for Lion-7B.}
\label{tab:ablation_r}
\end{table*}

\paragraph{The ratio $r$ of generated hard and easy instructions}
We change the ratio of generated hard instructions to generated easy instructions from 1:0 (all hard) to 0:1 (all easy) and investigate its impact on average performance across three datasets. It can be seen from Table \ref{tab:ablation_r} that higher ratios of hard to easy instructions generally lead to improved performance, with a balanced ratio of 1:1 yielding the highest average scores.

\subsection{The Learning Dynamics of Lion}
\begin{figure}[!tbp]
\centering
\includegraphics[width=0.9\linewidth]{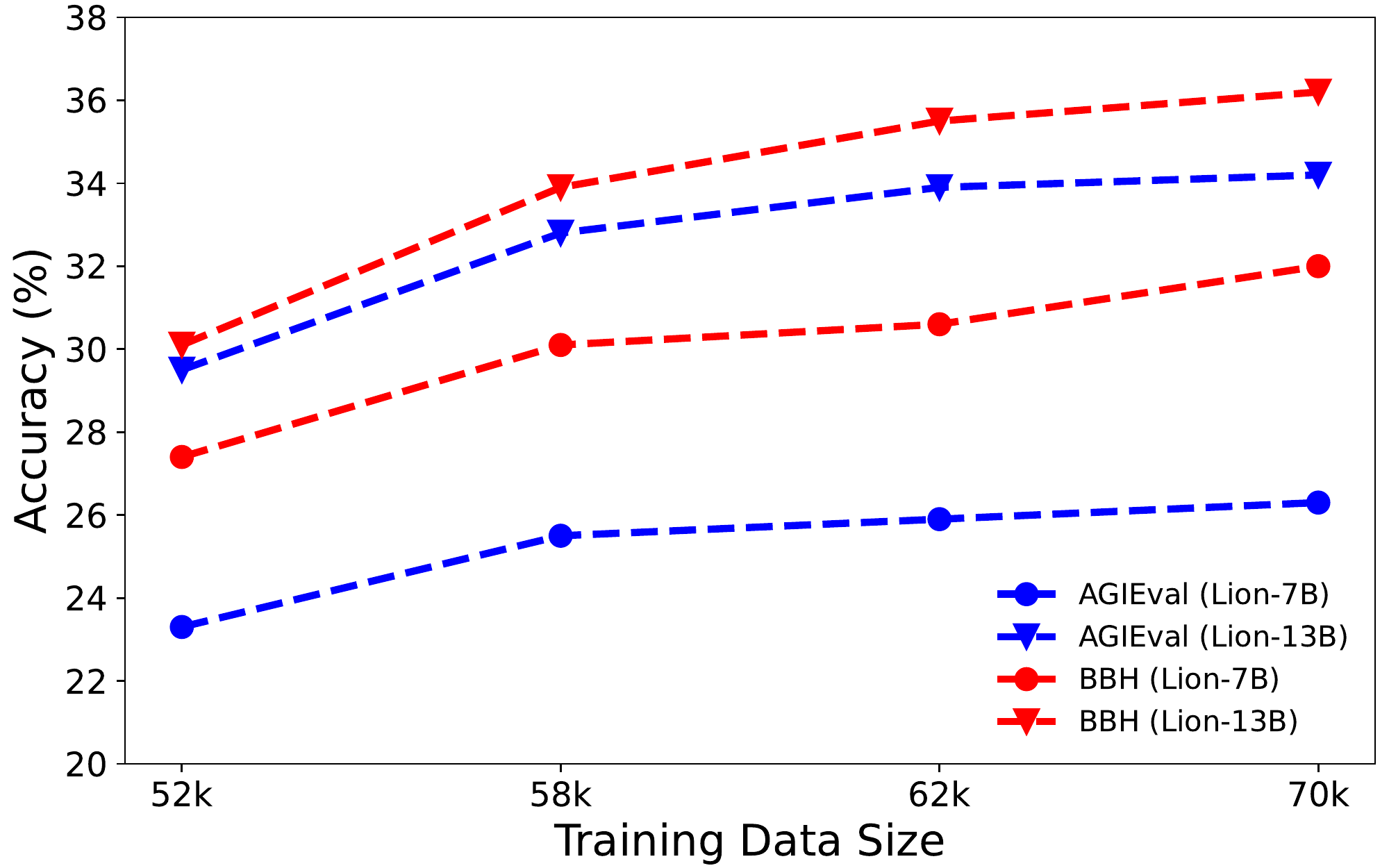}
\caption{
Performance of Lion-7B and Lion-13B on AGIEval and BBH through the training iterations.
}
\label{fig:learning_dynamics}
% \vspace{-1 cm}
\end{figure}

In Figure \ref{fig:learning_dynamics}, we delve into the learning dynamics of Lion by visualizing its performance on AGIEval and BBH throughout the training iterations.
The results clearly demonstrate that our adversarial knowledge distillation framework consistently enhances the performance of the student model as the iterations progress.
Notably, the most significant improvement in capability occurs in the first iteration, suggesting the usefulness of the identification of challenging example patterns (refer Figure \ref{fig:model2}).

\subsection{Case Studies}
To clearly compare the generated response quality between our model and other baselines, we provide nine case studies sampled from Vicuna-instruction, AGIEval, and BBH in Appendix \ref{sec:case_study}.
Table \ref{tab:case1} showcases the responses of various models to a math instruction.
It can be seen that only Lion and ChatGPT provide the correct answer and follow the correct problem-solving steps.
A counterfactual case is shown in Table \ref{tab:case2}, where ChatGPT provides a relevant answer that considers the potential impacts of Newton focusing on biology instead of physics, but it lacked details and depth. 
Lion, on the other hand, offered a more detailed and engaging response that explored different possibilities such as the development of biophysics or discovering new principles that could be applied to both fields. 
Lion's response also considered the potential implications of Newton's work on motion, force, gravity, and thermodynamics in biology, providing a more comprehensive answer.

% A counterfactual case is shown in Table \ref{tab:case2}, where ChatGPT provides a relevant answer that considers the potential impacts of Newton focusing on biology instead of physics, but it lacked details and depth. 
% Lion, on the other hand, offered a more detailed and engaging response that explored different possibilities such as the development of biophysics or discovering new principles that could be applied to both fields. 
% Lion's response also considered the potential implications of Newton's work on motion, force, gravity, and thermodynamics in biology, providing a more comprehensive answer.

\renewcommand{\dblfloatpagefraction}{.9}

%% file: 5_conclusion.tex
\section{Conclusion}
This paper presents an innovative adversarial knowledge distillation framework for distilling a proprietary LLM into a compact, open-source student model. 
While previous methodologies have concentrated on unidirectional knowledge transfer, our approach seeks to integrate ``feedback'' into the learning process. 
Leveraging the versatile role adaptability of LLMs, we prompt the proprietary model to identify ``hard'' instructions and generate new ``hard'' instructions for the student model, creating a three-stage adversarial loop of imitation, discrimination, and generation. 
This approach allows us to refine the student model's performance iteratively, efficiently bootstrapping its proficiency. 
% We apply this framework to the distillation of ChatGPT into a compact 7B-parameter model, Lion. Despite being trained on a modest 70k instruction-following data, Lion exhibits an impressive 95\% capability of ChatGPT, outperforming previous baselines in automatic and human evaluations. 
% This pioneering application of AKD to LLMs highlights the potential of iterative knowledge transfer, and we hope the release of the Lion model can promote continued exploration in this space.
% This innovative use of AKD in LLMs underscores the transformative potential of iterative knowledge transfer. We anticipate that the release of the Lion model will serve as a powerful catalyst for sustained exploration and advancement in this domain.
We aspire that our model, named Lion, may serve as a baseline to reflect the performance of ChatGPT, especially the open-source instruction-following language model baseline for our community.

%% file: 6_limitations.tex
\section*{Limitations and Discussions}

\paragraph{The Model Capability}
We have identified that Lion is subject to certain constraints:
1) A recent study \cite{DBLP:journals/corr/abs-2305-15717} asserts that ``model imitation is a false promise'' since imitation models are adept at mimicking ChatGPT’s style but fall short in improving LMs across more challenging tasks.
While Lion still lags behind its teacher model ChatGPT in handling intricate reasoning tasks (as shown in our experiments), it demonstrates promising improvements compared to previous imitation models.
Therefore, our adversarial knowledge distillation framework may provide a more effective way for knowledge transfer.
2) Since our training data doesn't encompass dialogues, Lion struggles to manage multi-turn conversations.
3) Due to computational resource constraints, Lion's maximum sequence length is limited to 1024. Consequently, it faces challenges when dealing with long documents.
% 4) The model hasn't been thoroughly optimized to ensure safety or to mitigate potential toxicity or bias.
Despite these limitations, we envision Lion serving as an accessible springboard for future research endeavors aimed at addressing these limitations.

\paragraph{The Training Process}
To train a single student model, we request the gpt-3.5-turbo API around 450k times,
a number that is roughly 70\% of the WizardLM's usage of 624k \cite{xu2023wizardlm}. Nonetheless, this utilization incurs a considerable expense, nearing \$900. 
In contrast to methods like Alpaca \cite{alpaca} and WizardLM \cite{xu2023wizardlm}, which only fine-tune the student model once, our adversarial knowledge distillation method employs iterative parametric updates to the student model. While this iterative approach inevitably leads to slower iteration speed, it offers additional benefits.
% Besides, our adversarial knowledge distillation iteratively updates the student model, rather than only one time fine-tuning in other methods like Alpaca \cite{alpaca} and WizardLM \cite{xu2023wizardlm}. This inevitably results in a slower iteration speed.
Finally, different from traditional adversarial knowledge distillation where the weights of the generator are iteratively updated, we use a black-box and parameter-frozen LLM (ChatGPT in our paper) to serve the role. Therefore, the quality of the LLM is quite essential in the generation of new instructions. 
% Moreover, our approach diverges from the conventional adversarial knowledge distillation methodology where the generator's weights are continually updated. Instead, we implement a black-box model with parameters set in stone, specifically a Language Learning Model (LLM), which in our study is represented by ChatGPT. This element underscores the vital role of the LLM's quality in the generation of new instructions.

% \paragraph{The Ablation Study}
% As previously mentioned, training a single student model can be a costly and time-consuming process. Therefore, we have not examined the impact of certain parameters in our proposed adversarial knowledge distillation framework, including the threshold $\tau$ for distinguishing between hard and easy instructions, as well as the ratio of generated hard and easy instructions. The exploration of parameter selection will be left for our future endeavors.

\paragraph{The Evaluation Metrics}
Though automated evaluations leveraging GPT-4 have showcased promising prospects in appraising chatbot performance, the technique is yet to reach a level of maturity and accuracy, especially considering the propensity of large language models to generate non-existent or ``hallucinated'' information. 
% Moreover, the absence of a unified standard for evaluating large language models diminishes the persuasiveness of human-based evaluations. 
Evaluating the efficacy of LLM across various tasks presents a considerable challenge since different tasks require quite different expertise \cite{wang2022selfinstruct}.
Therefore, the creation of a comprehensive, standardized evaluation system for chatbots is a prevailing research challenge that demands additional exploration and study.

%\jacky{Evluation Metrics}
%However, evaluating chatbots is never a simple task. While we only follow the proposed framework from the previous works shows a potential to automate chatbot assessment, it is not yet a rigorous approach.
%\jacky{Reasoning on Mathematics and Code Generation}
%We have noticed that, similar to other large language models, Vicuna has certain limitations. For instance, it is not good at tasks involving reasoning or mathematics, and it may have limitations in accurately identifying itself or ensuring the factual accuracy of its outputs. 

\section*{Ethics Statement}

\paragraph{Inherited Biases}
It is important to consider that the behavior of our distilled student models may exhibit potential toxicity, biases, or privacy issues~\cite{ray2023chatgpt, DBLP:journals/corr/abs-2304-05197} inherited from the larger teacher LLM. 
We anticipate that the advancements made in reducing anti-social behaviors in LLMs can also be utilized to enhance student language models.

\paragraph{License and Legality}
% Following Stanford Alpaca \cite{alpaca}, we have decided that in the future, the released weights of Lion are licensed for research use only. 
% Using the weights of Lion with LLaMA’s original weights is subject to Meta’s LLaMA License Agreement. 
% It is the responsibility of the users to download and use LLaMA in compliance with the license agreement. 
% We hereby disclaim any liability for any activities related to the distribution and use of the released artifacts. 
Based on Stanford Alpaca's guidelines \cite{alpaca}, we have determined that the weights of Lion will be exclusively licensed for research purposes in the future.
Utilizing Lion's weights alongside LLaMA's original weights must adhere to Meta's LLaMA License Agreement.
Users are responsible for acquiring and utilizing LLaMA in accordance with the license agreement.

\paragraph{Safety}
% Unlike ChatGPT \cite{openai2022chatgpt}, Lion does not rely on human feedback to suppress unwanted behaviors. Instead, Lion learns to avoid such behaviors by imitating ChatGPT.
% However, it is important to acknowledge that there are potential risks associated with the use of Lion for malicious purposes, especially when we release the
% weights in the future.
% For future work, we would like to utilize the technic of Reinforcement Learning from Human Feedback (RLHF)~\cite{ouyang2022training} to acquire a better access control.
% Besides, Meta has implemented an access application process that can help control the distribution of LLaMA models and minimize the potential risks associated with their use, which is also an alternative option.
% DBLP:journals/corr/abs-2304-05197
Unlike ChatGPT \cite{openai2022chatgpt}, Lion does not rely on human feedback to mitigate undesired behaviors. Instead, Lion learns to avoid such behaviors by imitating ChatGPT. However, it is important to acknowledge the potential risks associated with using Lion for malicious purposes, especially upon releasing its weights in the future.
For future work, we aim to incorporate the technique of Reinforcement Learning from Human Feedback (RLHF)~\cite{ouyang2022training} to enhance access control. Additionally, Meta has implemented an access application process that can help regulate the distribution of LLaMA models and minimize the potential risks associated with their usage, providing an alternative option.

\section*{Acknowledgements}
W.\ Wang was also affiliated with Guangzhou Municipal Key Laboratory of Materials Informatics, The Hong Kong University of Science and Technology (Guangzhou), China. He was supported by HKUST(GZ) Grant G0101000028, GZU-HKUST Joint Research Collaboration Grant GZU22EG04, CCF-HuaweiDBC202302, and Guangzhou Municipal Science and Technology Project (No.\ 2023A03J0003). 

% We sincerely thank our colleagues and friends who provided their private OpenAI APIKEY for our research usage. They are Ying Lin, Mingyu Yang, Zhihao Li, Bo Huang, Yinan Fan, Zheng Wei, Yuebin Ruan, Xinyi Wang, Siqian Shi, Yinhe Qi, Haotian Bai, Xing Liu and Liang Zhao.

%% file: 7_appendix.tex
\appendix

\section{Data Statistics}
\label{appendix:data stat}
Table \ref{tab:agieval_sat} and Table \ref{tab:bbheval_sat} show the data statistics of AGIEval and BIG-Bench Hard, respectively.

\begin{table}[!h]
\footnotesize
\centering
\begin{tabular}{l|c|c}
\toprule
\textbf{Task}          & \textbf{\# Examples} & \textbf{\# Choices} \\ \midrule
AQuA-RAT               & 254                  & 5                   \\
LogiQA                 & 651                  & 4                   \\
LSAT-AR                & 230                  & 5                   \\
LSAT-LR                & 510                  & 5                   \\
LSAT-RC                & 269                  & 5                   \\
SAT-Math               & 220                  & 4                   \\
SAT-English            & 206                  & 4                   \\
SAT-English (w/o Psg.) & 206                  & 4           
\\ \bottomrule
\end{tabular}
\caption{Statistics of AGIEval dataset.}
\label{tab:agieval_sat}
\end{table}

\begin{table}[!h]
\scriptsize
\centering
\begin{tabular}{l|c|c}
\toprule
\textbf{Task}                         & \textbf{\# Examples} & \textbf{\# Choices} \\ \midrule
Boolean Expressions                   & 250                  & 2                   \\
Causal Judgement                      & 187                  & 2                   \\
Date Understanding                    & 250                  & 6                   \\
Disambiguation QA                     & 250                  & 4                   \\
Formal Fallacies                      & 250                  & 2                   \\
Geometric Shapes                      & 250                  & 11                  \\
Hyperbaton                            & 250                  & 2                   \\
Logical Deduction (5 objects)         & 250                  & 5                   \\
Logical Deduction (7 objects)         & 250                  & 7                   \\
Logical Deduction (3 objects)         & 250                  & 3                   \\
Movie Recommendation                  & 250                  & 5                   \\
Navigate                              & 250                  & 2                   \\
Penguins in a Table                   & 146                  & 5                   \\
Reasoning about Colored Objects       & 250                  & 18                  \\
Ruin Names                            & 250                  & 11                   \\
Salient Translation Error Detection   & 250                  & 6                   \\
Snarks                                & 178                  & 2                   \\
Sports Understanding                  & 250                  & 2                   \\
Temporal Sequences                    & 250                  & 4                   \\
Tracking Shuffled Objects (5 objects) & 250                  & 5                   \\
Tracking Shuffled Objects (7 objects) & 250                  & 7                   \\
Tracking Shuffled Objects (3 objects) & 250                  & 3                   \\
Web of Lies                           & 250                  & 2 \\ \bottomrule                 
\end{tabular}
\caption{Statistics of BIG-Bench Hard dataset.}
\label{tab:bbheval_sat}
\end{table}

\section{Baselines}
\label{appendix:baseline}
\begin{itemize}
\item \textbf{LLaMA} \cite{touvron2023llama} is a collection of foundation language models ranging from 7B to 65B parameters. It is trained on trillions of tokens from publicly available datasets and is demonstrated to outperform larger-size LLMs such as GPT-3 (175B) across a multitude of benchmarks. We use the official code from LLaMA \footnote{\url{https://github.com/facebookresearch/llama}}.
\item \textbf{Alpaca} \cite{alpaca} is a project initiated by Stanford University with the objective of developing and disseminating an open-source model that adeptly follows instructions. It is based on LLaMA and fine-tuned on 52K instruction-following examples generated by querying OpenAI’s text-davinci-003 model. On the self-instruct evaluation set, Alpaca mirrors text-davinci-003, but is notably more compact and cost-effective to reproduce. We use the official code from Alpaca \footnote{\url{https://github.com/tatsu-lab/stanford_alpaca}}.
\item \textbf{WizardLM} \cite{xu2023wizardlm} employs LLMs instead of humans to automatically mass-produce open-domain instructions of various difficulty levels, to improve the performance of LLMs. 
It uses an Evol-Instruct method to bootstrap the 52k instruction-following examples of Alapca into a larger set of 250k more intricate instructions. Out of this larger set, 70k examples were selected to fine-tune LLaMA. We use WizardLM-7B-V1.0 from the official code \footnote{\url{https://github.com/nlpxucan/WizardLM}}.
\item \textbf{Vicuna} \cite{vicuna2023}, a superior open-source chatbot, excels in generating fluid and captivating responses to user queries. It is based on LLaMA and fine-tuned on 70K user-shared conversations collected from ShareGPT, a platform designed for sharing interactions with ChatGPT. Its impressive capabilities make it one of the leading open instruction-following models today. Vicuna achieves competitive performance against proprietary models such as ChatGPT and Bard \cite{Google2023}. We use Vicuna-7B-V1.1 and Vicuna-13B-V1.1 from FastChat \footnote{\url{https://github.com/lm-sys/FastChat}}.
\item \textbf{ChatGPT} \cite{openai2022chatgpt}, a product of OpenAI, is an advanced AI chatbot renowned for its ability to interact with users in an authentically human and engaging manner. The chatbot is built on powerful LLMs such as GPT-3.5 and GPT-4, which are trained on a vast corpus of internet text data.
ChatGPT undergoes fine-tuning via both supervised and reinforcement learning techniques, with the human trainers providing necessary feedback and direction.
\end{itemize}

\section{Implementation Details}
\paragraph{Training Hyperparameters}
The training process is conducted on 8 A100 GPUs.
During each iteration of adversarial knowledge distillation, the hyperparameters for training are shown in Table \ref{tab:hyperparameters}.

\begin{table}[h]
\small
\centering
\begin{tabular}{l|c|c}
\toprule
\textbf{Hyperparameter} & \textbf{Lion-7B} & \textbf{Lion-13B} \\ \midrule
Batch size              & 128              & 128               \\
Learning rate           & 2e-5             & 2e-5              \\
Epoches                 & 3                & 3                 \\
Max length              & 1024             & 1024              \\
Optimizer               & AdamW            & AdamW             \\
Scheduler               & cosine           & cosine            \\
Weight decay            & 0                & 0                 \\
Warmup ratio            & 0.03             & 0.03           \\ \bottomrule
\end{tabular}
\caption{\label{tab:hyperparameters}
Training hyperparameters.}
\end{table}

\label{sec:gpt_parameter}
\paragraph{Querying the gpt-3.5-turbo API}
We use different sets of hyperparameters when querying the gpt-3.5-turbo API for different roles (Teacher, Referee, Generator). These hyperparameters are found to work well and we listed them in Table \ref{tab:api}.

\begin{table}[h]
\tiny
\centering
\begin{tabular}{l|c|c|c|c}
\toprule
\textbf{Role}  & 
\textbf{temperature}  &
\textbf{top\_p}  &
\textbf{beam\_size (n)}  &
\textbf{max\_tokens}
\\
\midrule
Teacher   & 0.7 & 1.0  & 1 & 1024 \\
Referee   & 0.2 & 1.0  & 1 & 512 \\
Generator   & 1.0 & 1.0  & 1 & 512 \\
\bottomrule
\end{tabular}
\caption{\label{tab:api}
Hyperparameters for querying OpenAI gpt-3.5-turbo API under different roles.}
\end{table}

\section{Prompt Templates for Our Adversarial Distillation Framework}
\label{sec:prompt template}
Fine-tuning an LLM (i.e. ChatGPT) is costly and intricate, human-tailored prompt templates are utilized to solve various tasks~\cite{DBLP:conf/nips/Wei0SBIXCLZ22, 
DBLP:journals/corr/abs-2309-08303,
DBLP:conf/acl/ChanLCLSWS23,
DBLP:conf/emnlp/JiangZW22,
DBLP:conf/icmlc2/ChanC23}.
The prompt template of the \textbf{Teacher} for generating responses is shown in Table \ref{tab:response}.
The prompt template of the \textbf{Referee} for comparing the quality of two responses generated by two AI assistants is shown in Table \ref{tab:discriminator_gpt-3.5-turbo}.
The prompt templates of the \textbf{Generator} for generating new hard instructions and new easy instructions are shown in Table \ref{tab:generator_hard} and Table \ref{tab:generator_no_hard}, respectively.

\begin{table*}[!h]
\small
\centering
\begin{tabular}{l|l}
\toprule
system content &
\parbox[c]{13cm}{
\texttt{You are a helpful assistant that generates a response to a given task instruction.}
}\\
\toprule
{user content} &
% \parbox[c]{13cm}{
% (if $input$ is not None) \\
% % \texttt{Below is an instruction that describes a task, paired with an input that provides further context. Write a response that appropriately completes the request.} \\ \\
% \texttt{\#\#\# Instruction:} \\ 
% $instruction$ \\ \\
% \texttt{\#\#\# Input:} \\ 
% $input$ \\ \\
% \texttt{\#\#\# Response:} \\
% }
% \\ \cline{2-2}
% \addlinespace[1ex]
% &
\parbox[c]{13cm}{
% (if $input$ is None) \\
% \texttt{Below is an instruction that describes a task. Write a response that appropriately completes the request.} \\ \\
\texttt{\#\#\# Instruction:} \\ 
\texttt{\{instruction\}} \\ \\
\texttt{\#\#\# Response:} \\
} \\
\bottomrule
\end{tabular}
\caption{Prompt template of gpt-3.5-turbo for generating responses. Note that the original instruction in Alpaca is composed of an instruction
prompt and an instance input. For example, the instruction prompt is ``write an abstract about the following method'', and the instance input is ``knowledge distillation''.
For a better adaption to real-world scenarios, we concatenate the instruction prompt and the instruction prompt into one instruction using a line break.}
\label{tab:response}
\end{table*}

% \section{Prompt Format for AGIEval and BBH}
% \label{sec:prompt reasoning}

\section{Case Studies}
\label{sec:case_study}
Here we show 3 cases in Table \ref{tab:case1}, \ref{tab:case2}, and \ref{tab:case3} to clearly compare the open-ended generation performance among various models including our Lion-13B, LLaMA-13B, Alpaca-13B, Vicuna-13B, and ChatGPT.

Besides, we show 6 cases in Table \ref{tab:case3}, \ref{tab:case4}, \ref{tab:case5}, \ref{tab:case6}, \ref{tab:case7}, and \ref{tab:case8} to clearly compare the reasoning capability among various models including our Lion-13B, Vicuna-13B, and ChatGPT.
We utilize \textcolor{green}{\Large\textbf{\ding{51}}} and \textcolor{red}{\Large\textbf{\ding{55}}} to denote whether the response is correct or incorrect, respectively.

\begin{table*}[!h]
\small
\centering
\begin{tabular}{l|l}
\toprule
system content &
\parbox[c]{13cm}{
\texttt{You are a helpful and precise assistant for checking the quality of the answer.}
}\\
\toprule
user content &
\parbox[c]{13cm}{
$\left[ \texttt{Instruction} \right]$\\ 
\texttt{\{instruction\}} \\ \\ 
$\left[ \texttt{The Start of Assistant 1's Answer} \right]$\\ 
\texttt{\{answer\_1\}}\\ 
$\left[ \texttt{The End of Assistant 1's Answer} \right]$\\ \\
$\left[ \texttt{The Start of Assistant 2's Answer} \right]$\\ 
\texttt{\{answer\_2\}} \\
$\left[ \texttt{The End of Assistant 2's Answer} \right]$\\ \\
$\left[ \texttt{System} \right]$\\ 
\texttt{We would like to request your feedback on the performance of two AI assistants in response to the user instruction and input displayed above.} \\ \\
\texttt{Please rate the helpfulness, relevance, accuracy, and level of detail of their responses. Each assistant receives an overall score on a scale of 1 to 10, where a higher score indicates better overall performance.} \\ \\
\texttt{Please first provide a comprehensive explanation of your evaluation, avoiding any potential bias and ensuring that the order in which the responses were presented does not affect your judgment. Then, output two lines indicating the scores for Assistant 1 and 2, respectively.} \\ \\
\texttt{Output with the following format:} \\
\texttt{Evaluation evidence: <your evaluation explanation here>} \\
\texttt{Score of the Assistant 1: <score>} \\
\texttt{Score of the Assistant 2: <score>}
}
\\
\bottomrule
\end{tabular}
\caption{Prompt template of gpt-3.5-turbo for comparing the quality of two responses generated by two AI assistants.}
\label{tab:discriminator_gpt-3.5-turbo}
\end{table*}

% \begin{table*}[!h]
% \small
% \centering
% \begin{tabular}{l|l}
% \toprule
% system content &
% \parbox[c]{13cm}{
% \texttt{You are a helpful and precise assistant for checking the quality of the answer.}
% }\\
% \toprule
% user content &
% \parbox[c]{13cm}{
% $\left[ \texttt{Question} \right]$\\ 
% $question$\\ \\ 
% $\left[ \texttt{The Start of Assistant 1's Answer} \right]$\\ 
% $answer_1$\\ \\ 
% $\left[ \texttt{The End of Assistant 1's Answer} \right]$\\ \\
% $\left[ \texttt{The Start of Assistant 2's Answer} \right]$\\ 
% $answer_2$\\ \\ 
% $\left[ \texttt{The End of Assistant 2's Answer} \right]$\\ \\
% $\left[ \texttt{System} \right]$\\ 
% \texttt{We would like to request your feedback on the performance of two AI assistants in response to the user question displayed above.} \\
% \texttt{Please rate the helpfulness, relevance, accuracy, level of details of their responses. Each assistant receives an overall score on a scale of 1 to 10, where a higher score indicates better overall performance.} \\
% \texttt{Please first output a single line containing only two values indicating the scores for Assistant 1 and 2, respectively. The two scores are separated by a space. In the subsequent line, please provide a comprehensive explanation of your evaluation, avoiding any potential bias and ensuring that the order in which the responses were presented does not affect your judgment.} \\ \\
% }
% \\
% \bottomrule
% \end{tabular}
% \caption{Prompt template of gpt-4 for comparing the quality of two responses generated by two AI assistants.}
% \label{tab:evaluator_gpt-4}
% \end{table*}

\begin{table*}[!h]
\small
\centering
\begin{tabular}{l|l}
\toprule
system content &
\parbox[c]{13cm}{
\texttt{You are a helpful assistant.}
}\\
\toprule
user content &
\parbox[c]{13cm}{
\texttt{I want you to act as an Instruction Creator.} \\ 
\texttt{Your goal is to draw inspiration from the \#Given Instruction\# to create a brand new instruction.} \\ 
\texttt{This new instruction should belong to the same domain and the same task type as the \#Given Instruction\#.} \\ 
\texttt{The LENGTH and difficulty level of the \#Created Instruction\# should be similar to that of the \#Given Instruction\#.} \\ 
\texttt{The \#Created Instruction\# must be reasonable and must be understood and responded to by humans.} \\
\texttt{'\#Given Instruction\#', '\#Created Instruction\#', 'given instruction' and 'created instruction' are not allowed to appear in \#Created Instruction\#.} \\ \\
\texttt{\#Given Instruction\#:} \\
\texttt{\{instruction\}} \\ \\
\texttt{\#Created Instruction\#:} \\
} \\
\bottomrule
\end{tabular}
\caption{Prompt template of gpt-3.5-turbo for generating new hard instructions.}
\label{tab:generator_hard}
\end{table*}

\begin{table*}[!h]
\small
\centering
\begin{tabular}{l|l}
\toprule
system content &
\parbox[c]{13cm}{
\texttt{You are a helpful assistant.}
}\\
\toprule
user content &
\parbox[c]{13cm}{
\texttt{I want you to act as an Instruction Creator.} \\ 
\texttt{Your goal is to draw inspiration from the \#Given Instruction\# to create a brand new instruction.} \\ 
\texttt{This new instruction should belong to the same domain as the \#Given Instruction\# but be even more rare.} \\ 
\texttt{The LENGTH and difficulty level of the \#Created Instruction\# should be similar to that of the \#Given Instruction\#.} \\ 
\texttt{The \#Created Instruction\# must be reasonable and must be understood and responded to by humans.} \\
\texttt{'\#Given Instruction\#', '\#Created Instruction\#', 'given instruction' and 'created instruction' are not allowed to appear in \#Created Instruction\#.} \\ \\
\texttt{\#Given Instruction\#:} \\
\texttt{\{instruction\}} \\ \\
\texttt{\#Created Instruction\#:} \\
} \\
\bottomrule
\end{tabular}
\caption{Prompt template of gpt-3.5-turbo for generating new easy instructions.}
\label{tab:generator_no_hard}
\end{table*}

\begin{table*}[!h]
\small
\centering
\renewcommand{\arraystretch}{1}
\begin{tabular}{ll}
\toprule
\multicolumn{2}{l}{
\parbox[c]{13cm}{
\textbf{Instruction:} If the endpoints of a line segment are $(2,-2)$ and $(10,4)$, what is the length of the segment?
}
}\\
\toprule
\multicolumn{1}{l|}{\textbf{Lion}}
&
\parbox[c]{13cm}{
To find the length of the line segment with endpoints $(2,-2)$ and $(10,4)$, we can use the distance formula. The distance formula is derived from the Pythagorean theorem and is given as:
\\ \\
$d = \sqrt{((x2 - x1)^2 + (y2 - y1)^2)}$
\\ \\
where $(x1, y1)$ and $(x2, y2)$ are the coordinates of the endpoints of the line segment.
\\ \\
Substituting the given values, we get:
\\ \\
$d = \sqrt{((10 - 2)^2 + (4 - (-2))^2)}$
\\ \\
Simplifying this expression, we get:
\\ \\
$d = \sqrt{((8)^2 + 6^2)}$
\\ \\
$d = \sqrt{(64 + 36)}$
\\ \\
$d = \sqrt{(100)}$
\\ \\
$d = 10$
\\ \\
Therefore, the length of the line segment with endpoints $(2,-2)$ and $(10,4)$ is $10$ units. \textcolor{green}{\Large\textbf{\ding{51}}}
} 
\\ \midrule

\multicolumn{1}{l|}{\textbf{LLaMA}}
&
\parbox[c]{13cm}{
The length of the line segment is $10$ units. \textcolor{green}{\Large\textbf{\ding{51}}}
} 
\\ \midrule

\multicolumn{1}{l|}{\textbf{Alpaca}}
&
\parbox[c]{13cm}{
The length of the line segment is $8$ units. \textcolor{red}{\Large\textbf{\ding{55}}}
} 
\\ \midrule

\multicolumn{1}{l|}{\textbf{Vicuna}}
&
\parbox[c]{13cm}{
The length of the line segment is the distance between the two points. To find the length, you can use the Pythagorean theorem, which states that the sum of the squares of the distances from a point to the two endpoints is equal to the square of the distance between the two endpoints.
\\ \\
In this case, the distance from the point $(2, -2)$ to the point $(10, 4)$ is given by the formula:
\\ \\
$\sqrt{((x2 - x1)^2 + (y2 - y1)^2)}$ \\
$= \sqrt{((10 - 2)^2 + (4 - (-2))^2)}$ \\
$= \sqrt{(18^2 + 4^2)}$ \\
$= \sqrt{(324 + 16)}$ \\
$= \sqrt{(340)}$ \\
$= 17.973$
\\ \\
So the length of the line segment is $17.973$. \textcolor{red}{\Large\textbf{\ding{55}}}
} 
\\ \midrule

% \multicolumn{1}{l|}{\textbf{WizardLM}}
% &
% \parbox[c]{13cm}{
% To find the length of the line segment, we need to use the distance formula:
% $d = \sqrt{((x2 - x1)^2 + (y2 - y1)^2)}$ \\
% where $(x1, y1)$ and $(x2, y2)$ are the coordinates of the two endpoints, and $d$ is the length of the segment. \\
% Plugging in the values, we get: \\
% $d = \sqrt{((10 - 2)^2 + (4 - (-2))^2)}$ \\
% $d = \sqrt{(16 + 8)}$ \\
% $d = \sqrt{(24)}$ \\
% $d = 4 * \sqrt{(3)}$ \\
% Therefore, the length of the line segment is $4 * \sqrt{(3)}$.
% } 
% \\ \midrule

\multicolumn{1}{l|}{\textbf{ChatGPT}}
&
\parbox[c]{13cm}{
To find the length of a line segment with two endpoints $(x1, y1)$ and $(x2, y2)$, we use the distance formula:
\\ \\
$distance = \sqrt{((x2 - x1)^2 + (y2 - y1)^2)}$
\\ \\
Plugging in the coordinates of the given endpoints, we get:
\\ \\
$distance = \sqrt{((10 - 2)^2 + (4 - (-2))^2)}$ \\
$distance = \sqrt{(8^2 + 6^2)}$ \\
$distance = \sqrt{(64 + 36)}$ \\
$distance = \sqrt{(100)}$ \\
$distance = 10$ \\
\\ \\
Therefore, the length of the line segment is $10$ units. \textcolor{green}{\Large\textbf{\ding{51}}}
} 
\\

\bottomrule
\end{tabular}
\caption{A math case from Vicuna-Instructions.}
\label{tab:case1}
\renewcommand{\arraystretch}{1}
\end{table*}

\begin{table*}[!h]
\small
\centering
\renewcommand{\arraystretch}{1}
\begin{tabular}{ll}
\toprule
\multicolumn{2}{l}{
\parbox[c]{13cm}{
\textbf{Instruction:} What if Isaac Newton had focused on biology instead of physics?
}
}\\
\toprule
\multicolumn{1}{l|}{\textbf{Lion}}
&
\parbox[c]{13cm}{
Isaac Newton is renowned for his contributions to physics, particularly his laws of motion and universal gravitation. However, what if Newton had instead chosen to focus on biology? \\ \\Newton's work in physics was heavily influenced by his fascination with the natural world and his desire to understand how it worked. Similarly, had he chosen to study biology, he may have made significant contributions to the field. Newton's mathematical abilities and his interest in the natural world would have made him a formidable biologist.\\ \\One area where Newton may have made a significant impact in biology is in the field of taxonomy. Newton was a keen observer of nature and had a particular interest in plants. He spent much of his time studying and cataloguing the various species of plants around him. He even developed a system for classifying plants based on their physical characteristics.\\ \\In addition to his work in taxonomy, Newton may have also made important contributions to the field of physiology. He was interested in the workings of the human body and may have studied the mechanics of muscles and the circulatory system.\\ \\Overall, while Isaac Newton is most well-known for his work in physics, his interests and abilities would have made him a formidable biologist as well. Who knows what discoveries he might have made had he chosen to focus on the natural world in a different way.
} 
\\ \midrule

\multicolumn{1}{l|}{\textbf{LLaMA}}
&
\parbox[c]{13cm}{
I'm not sure. If he did so, he wouldn't have discovered the law of gravity.
} 
\\ \midrule

\multicolumn{1}{l|}{\textbf{Alpaca}}
&
\parbox[c]{13cm}{
If Isaac Newton had focused on biology instead of physics, he might have made important contributions to the field of biology, such as developing a scientific theory of evolution or explaining the processes of cell growth and development. He might also have studied and discovered new principles of genetics and developed methods to study and treat diseases.
} 
\\ \midrule

\multicolumn{1}{l|}{\textbf{Vicuna}}
&
\parbox[c]{13cm}{
Isaac Newton is widely regarded as one of the most influential scientists in history for his work in physics, particularly for his development of the laws of motion and universal gravitation. However, it is worth noting that Newton also made significant contributions to other fields, including mathematics, optics, and natural philosophy.\\ \\If Newton had decided to focus on biology instead of physics, it is possible that he would have made significant contributions to the field. However, it is difficult to say exactly what those contributions would have been without knowing more about his specific interests and areas of focus.\\ \\Newton was a prolific scientist and wrote extensively on a wide range of topics, including natural philosophy, mathematics, and theology. He also conducted extensive research in optics, including the study of light and color, and made important contributions to the field of mathematics, including the development of calculus.\\ \\It is possible that if Newton had focused on biology, he would have made significant contributions to the understanding of living organisms and their behavior. However, it is also possible that his work in physics and mathematics would have had a greater impact on the world.
} 
\\ \midrule

% \multicolumn{1}{l|}{\textbf{WizardLM}}
% &
% \parbox[c]{13cm}{
% If Isaac Newton had focused on biology instead of physics, it is possible that he would have made significant contributions to the field of biology. However, it is also possible that he may not have made as many groundbreaking discoveries in physics as he did.
% \\ \\
% Newton's work in physics laid the foundation for many modern day scientific discoveries and advancements. His laws of motion and laws of universal gravitation are still studied and applied today in fields such as engineering, physics, and space exploration.
% \\ \\
% In biology, Newton may have focused on areas such as the behavior of living organisms, the study of evolution, and the classification of species. However, it is difficult to predict exactly what discoveries he may have made in this field as his interests and focus may have been different.
% \\ \\
% Overall, it is difficult to say for certain what path Isaac Newton's life would have taken had he focused on biology instead of physics. While he may have made significant contributions to the field of biology, it is possible that his legacy in physics would have been different.
% } 
% \\ \midrule

\multicolumn{1}{l|}{\textbf{ChatGPT}}
&
\parbox[c]{13cm}{
If Isaac Newton had focused on biology instead of physics, the scientific world may be vastly different today. His advancements in mathematics, such as calculus, would have certainly been useful in the field of biology. Newton's curiosity, analytical mind, and dedication to scientific inquiry would have been equally effective in biology as they were in physics. He may have made significant contributions to the field of genetics or evolutionary theory.
\\ \\
On the other hand, his contributions to physics were so groundbreaking that the subject may have been hindered without his contributions. It's difficult to say exactly what would have happened, but it's likely that the course of scientific history would have been different.
} 
\\

\bottomrule
\end{tabular}
\caption{A counterfactual case from Vicuna-Instructions.}
\label{tab:case2}
\renewcommand{\arraystretch}{1}
\end{table*}

\begin{table*}[!ht]
\scriptsize
\centering
\renewcommand{\arraystretch}{1}
\begin{tabular}{ll}
\toprule
\multicolumn{2}{l}{
\parbox[c]{13cm}{
\textbf{Instruction:} How can I develop my critical thinking skills?
}
}\\
\toprule
\multicolumn{1}{l|}{\textbf{Lion}}
&
\parbox[c]{13cm}{
Developing critical thinking skills is an important aspect of personal growth and professional development. Critical thinking enables us to analyze and evaluate information, ideas, and arguments objectively, make informed decisions, and solve complex problems. Here are some strategies that can help you develop your critical thinking skills:\\ \\1. Read widely and deeply: Reading widely exposes you to different perspectives and ideas, while reading deeply helps you to understand and analyze complex concepts. Choose books that are challenging and outside your comfort zone, and take notes as you read.\\ \\2. Ask questions: Asking questions is an essential part of critical thinking. Start by asking \"why\" and \"how\" questions to gain a deeper understanding of a topic. Then, ask \"what if\" questions to explore possible outcomes and consequences.\\ \\3. Practice analysis and evaluation: Critical thinking involves analyzing and evaluating information and arguments. Start by breaking down complex ideas into smaller parts, and then evaluate each part based on evidence, reasoning, and credibility.\\ \\4. Seek diverse perspectives: Surround yourself with people who have different backgrounds, experiences, and perspectives. Listen to their ideas and opinions, and challenge your own assumptions and biases.\\ \\5. Practice decision-making: Critical thinking is essential in decision-making. Practice making decisions by weighing the pros and cons, considering potential outcomes, and seeking advice from others.\\ \\6. Take courses and workshops: Enrolling in courses and workshops that focus on critical thinking, problem-solving, and decision-making can provide you with the tools and techniques to develop your critical thinking skills.\\ \\By implementing these strategies, you can develop your critical thinking skills and become better equipped to analyze, evaluate, and make informed decisions in your personal and professional life.
} 
\\ \midrule

\multicolumn{1}{l|}{\textbf{LLaMA}}
&
\parbox[c]{13cm}{
Critical thinking is a vital skill to have, but it can be difficult to develop. Here are some ways you can start developing your critical thinking skills:\\
1. Analyze and interpret: When you're presented with information, try to analyze and interpret it on your own terms. When you start to analyze and interpret information on your own, it will become easier to do so over time.\\
2. Assess your information: When you receive new information, try to assess whether it's valid or reliable. This is important for evaluating and forming your own opinions.\\
3. Use your intuition: Use your intuition to complement your critical thinking skills. When you pay attention to your gut feelings, they'll usually tell you when you need to double-check your information.\\
4. Learn from your mistakes: When you make a mistake, try not to beat yourself up about it. Instead, learn from your mistakes and use them to your advantage.
} 
\\ \midrule

\multicolumn{1}{l|}{\textbf{Alpaca}}
&
\parbox[c]{13cm}{
Developing critical thinking skills involves actively questioning assumptions, weighing evidence, and thinking analytically. It also involves being open-minded and considering various perspectives, as well as thinking logically and systematically. Practicing problem-solving and creative thinking, working through scenarios and dilemmas, and engaging in thoughtful conversations are all ways to improve critical thinking. Additionally, reading, reflecting, and writing can help increase your capacity for critical thinking.
} 
\\ \midrule

\multicolumn{1}{l|}{\textbf{Vicuna}}
&
\parbox[c]{13cm}{
Developing your critical thinking skills can help you to better analyze information, make more informed decisions, and solve complex problems. Here are some tips for improving your critical thinking skills:\\1. Practice active listening: Pay attention to what others are saying and try to understand their perspective.\\2. Ask questions: Be curious and ask questions to clarify information and better understand a situation.\\3. Analyze information: Take time to carefully analyze information and consider different viewpoints.\\4. Seek out diverse viewpoints: Expose yourself to a variety of viewpoints and ideas to challenge your own assumptions and biases.\\5. Practice logic and reasoning: Use logic and reasoning to evaluate arguments and make decisions.\\6. Learn from your mistakes: Be open to learning from your mistakes and using them as opportunities for growth.\\7. Read and watch a variety of sources: Read and watch a variety of sources to expand your knowledge and expose yourself to different perspectives.\\8. Engage in debates and discussions: Engage in debates and discussions with others to practice critical thinking and develop your skills.\\9. Take breaks and reflect: Take breaks and reflect on your thinking process to improve your critical thinking skills.\\10. Practice mindfulness: Mindfulness can help you to be more aware of your thoughts and biases, which can help you to think more critically.
} 
\\ \midrule

% \multicolumn{1}{l|}{\textbf{WizardLM}}
% &
% \parbox[c]{13cm}{
% As an AI language model, I do not have personal experiences, but here are some tips to help you develop critical thinking skills:
% \\ \\
% 1. Read widely: Expose yourself to diverse perspectives, ideas, and opinions by reading books, articles, and news from different sources.
% \\ \\
% 2. Question your assumptions: Always ask questions and challenge your own beliefs and assumptions.
% \\ \\
% 3. Analyze arguments: Analyze the reasoning behind an argument, not just the conclusion. Determine if the argument is logically sound and consider the potential biases.
% \\ \\
% 4. Think critically: Avoid jumping to conclusions or accepting things at face value. Instead, analyze information and consider alternative explanations.
% \\ \\
% 5. Seek feedback: Ask for feedback on your thinking and arguments from others. This can help identify blind spots and areas for improvement.
% \\ \\
% 6. Reflect on your thinking: Reflect on your thinking process and how it influences your decision-making. This can help you identify patterns and improve your critical thinking skills.
% \\ \\
% 7. Practice, practice, practice: Critical thinking skills are like any other skill, and they require practice to develop. Consistent effort and dedication can help you improve your critical thinking abilities.
% } 
% \\ \midrule

\multicolumn{1}{l|}{\textbf{ChatGPT}}
&
\parbox[c]{13cm}{
Sure, here are a few practical tips to develop critical thinking skills:

1. Ask questions: Don't simply accept information at face value. Ask questions that allow you to evaluate the reason, perspective and logic behind the information.
\\ \\
2. Analyze: Take the time to analyze and evaluate the information that you receive. This will help you to identify assumptions, biases, strengths and weaknesses of an argument.
\\ \\
3. Consider different perspectives: When analyzing information, try to consider different perspectives to gain a wider understanding of any topic.
\\ \\
4. Research and read extensively: Read a wide range of sources to get a better understanding of different viewpoints and opinions.
\\ \\
5. Practice logic and reasoning: Practice thinking through problems logically by breaking them down into smaller parts.
\\ \\
6. Establish a point of view: Consider developing an opinion or thesis statement, and then collect evidence to support that position.
\\ \\
7. Reflect on your own thinking: Consider how you arrived at a conclusion, and whether you have any biases or assumptions that may have influenced your thinking.
\\ \\
With time, and practicing these skills consistently, you can become a better critical thinker.
} 
\\

\bottomrule
\end{tabular}
\caption{A generic case from Vicuna-Instructions.}
\label{tab:case3}
\renewcommand{\arraystretch}{1}
\end{table*}

\begin{table*}[!ht]
\small
\centering
\renewcommand{\arraystretch}{1}
\begin{tabular}{ll}
\toprule
\multicolumn{2}{l}{
\parbox[c]{14.3cm}{
\textbf{Instruction:} Q: If 120 is reduced to 96, what is the reduction percent? Answer Choices: (A)30\% (B)40\% (C)20\% (D)10\% (E)5\% A: Among A through E, the answer is
}
}\\
\toprule
\multicolumn{1}{l|}{\textbf{Lion}}
&
\parbox[c]{13cm}{
To calculate the reduction percent, we need to divide the difference between the original number and the reduced number by the original number and then multiply by 100. \\ \\Here, the original number is 120 and the reduced number is 96. So, the difference between the two numbers is 120 - 96 = 24. \\ \\To calculate the percentage reduction, we divide the difference by the original number and then multiply by 100:\\ \\(24 / 120) x 100 = 20\%\\ \\Therefore, the correct answer is option (C) 20\%. \\ \\Option (A) 30\%, option (B) 40\%, option (D) 10\%, and option (E) 5\% are all incorrect. \textcolor{green}{\Large\textbf{\ding{51}}}
} 
\\ \midrule

\multicolumn{1}{l|}{\textbf{Vicuna}}
&
\parbox[c]{13cm}{
A. 30\% \textcolor{red}{\Large\textbf{\ding{55}}}
} 
\\ \midrule

\multicolumn{1}{l|}{\textbf{ChatGPT}}
&
\parbox[c]{13cm}{
To calculate the reduction percentage, we need to find the difference between the original value and the reduced value, and then divide it by the original value.
\\ \\
The difference between 120 and 96 is 120 - 96 = 24.
\\ \\
Now, divide the difference by the original value and multiply by 100 to get the percentage reduction:
\\ \\
(24 / 120) * 100 = 20\%
\\ \\
Therefore, the correct answer is (C) 20\%.  \textcolor{green}{\Large\textbf{\ding{51}}}
} 
\\

\bottomrule
\end{tabular}
\caption{An AQuA-RAT case from AGIEval.}
\label{tab:case4}
\renewcommand{\arraystretch}{1}
\end{table*}

\begin{table*}[!ht]
\small
\centering
\renewcommand{\arraystretch}{1}
\begin{tabular}{ll}
\toprule
\multicolumn{2}{l}{
\parbox[c]{14.3cm}{
\textbf{Instruction:} Q: not ( True ) and ( True ) is? Answer Choices: (A) True (B) False A: Among A through B, the answer is
}
}\\
\toprule
\multicolumn{1}{l|}{\textbf{Lion}}
&
\parbox[c]{13cm}{
The expression "not (True) and (True)" can be simplified as "(not True) and (True)" or "False and True". The "and" operator has a higher precedence than the "not" operator, so the first expression is evaluated as "False and True", which is false. 
\\ \\
Therefore, the correct answer is option (B) False. \textcolor{green}{\Large\textbf{\ding{51}}}
} 
\\ \midrule

\multicolumn{1}{l|}{\textbf{Vicuna}}
&
\parbox[c]{13cm}{
(B) False
\\ \\
\#\#\# Explanation:
\\ \\
This statement is false.
\\ \\
If the given statement is "not (True)" and "True", it would be equivalent to "not (True) and True", which would be true.
\\ \\
(This is because negation of "True" is "False", and "True" and "False" evaluate to "True".)
\\ \\
Therefore, the correct answer is (A) True. \textcolor{red}{\Large\textbf{\ding{55}}}
} 
\\ \midrule

\multicolumn{1}{l|}{\textbf{ChatGPT}}
&
\parbox[c]{13cm}{
(A) True. \textcolor{red}{\Large\textbf{\ding{55}}}
} 
\\

\bottomrule
\end{tabular}
\caption{A boolean expressions case from BIGBench Hard.}
\label{tab:case5}
\renewcommand{\arraystretch}{1}
\end{table*}

\begin{table*}[!ht]
\small
\centering
\renewcommand{\arraystretch}{1}
\begin{tabular}{ll}
\toprule
\multicolumn{2}{l}{
\parbox[c]{14.3cm}{
\textbf{Instruction:} Q: If you follow these instructions, do you return to the starting point? Take 8 steps. Turn around. Take 8 steps. Answer Choices: (A) Yes (B) No A: Among A through B, the answer is
}
}\\
\toprule
\multicolumn{1}{l|}{\textbf{Lion}}
&
\parbox[c]{13cm}{
According to the instructions given, if you follow them, you will return to the starting point. Therefore, the correct answer is Option A, "Yes". By taking 8 steps and turning around, you will end up back where you started. It is important to follow instructions carefully to achieve the desired outcome. In this case, following these instructions will ensure that you return to the starting point. \textcolor{green}{\Large\textbf{\ding{51}}}
} 
\\ \midrule

\multicolumn{1}{l|}{\textbf{Vicuna}}
&
\parbox[c]{13cm}{
(A) Yes \textcolor{green}{\Large\textbf{\ding{51}}}
} 
\\ \midrule

\multicolumn{1}{l|}{\textbf{ChatGPT}}
&
\parbox[c]{13cm}{
(B) No \textcolor{red}{\Large\textbf{\ding{55}}}
} 
\\

\bottomrule
\end{tabular}
\caption{A navigate case from BIGBench Hard.}
\label{tab:case6}
\renewcommand{\arraystretch}{1}
\end{table*}

\begin{table*}[!ht]
\small
\centering
\renewcommand{\arraystretch}{1}
\begin{tabular}{ll}
\toprule
\multicolumn{2}{l}{
\parbox[c]{14.3cm}{
\textbf{Instruction:} Q: Today is Christmas Eve of 1937. What is the date tomorrow in MM/DD/YYYY? Answer Choices: (A) 12/11/1937 (B) 12/25/1937 (C) 01/04/1938 (D) 12/04/1937 (E) 12/25/2006 (F) 07/25/1937 A: Among A through F, the answer is
}
}\\
\toprule
\multicolumn{1}{l|}{\textbf{Lion}}
&
\parbox[c]{13cm}{
The correct answer is (B) 12/25/1937.
\\ \\
To arrive at this answer, we need to consider the fact that Christmas Eve is the evening before Christmas Day. Therefore, the date of Christmas Day is one day after Christmas Eve. In this case, the date of Christmas Eve is 12/24/1937, and the date of Christmas Day is 12/25/1937. Therefore, the date tomorrow in MM/DD/YYYY format is 12/25/1937. 
\\ \\
Option (A) is incorrect because it gives the date as 12/11/1937, which is incorrect. 
\\ \\
Option (C) is incorrect because it gives the date as 01/04/1938, which is incorrect.
\\ \\
Option (D) is incorrect because it gives the date as 12/04/1937, which is incorrect.
\\ \\
Option (E) is incorrect because it gives the date as 12/25/2006, which is incorrect.
\\ \\
Option (F) is incorrect because it gives the date as 07/25/1937, which is incorrect. \textcolor{green}{\Large\textbf{\ding{51}}}
} 
\\ \midrule

\multicolumn{1}{l|}{\textbf{Vicuna}}
&
\parbox[c]{13cm}{
(A) 12/11/1937

\#\#\# Explanation:

The date tomorrow, on Christmas Eve of 1937, would be 12/11/1937. \textcolor{red}{\Large\textbf{\ding{55}}}
% \textcolor{red}{\textbf{$\times$}}

} 
\\ \midrule

\multicolumn{1}{l|}{\textbf{ChatGPT}}
&
\parbox[c]{13cm}{
(C) 01/04/1938. \textcolor{red}{\Large\textbf{\ding{55}}}
} 
\\

\bottomrule
\end{tabular}
\caption{A data understanding case from BIGBench Hard.}
\label{tab:case7}
\renewcommand{\arraystretch}{1}
\end{table*}

\begin{table*}[!ht]
\small
\centering
\renewcommand{\arraystretch}{1}
\begin{tabular}{ll}
\toprule
\multicolumn{2}{l}{
\parbox[c]{14.3cm}{
\textbf{Instruction:} Q: Which statement is sarcastic? Answer Choices: (A) Hey just be happy then you won't be depressed anymore (B) Hey just be happy that you won't be depressed anymore A: Among A through B, the answer is
}
}\\
\toprule
\multicolumn{1}{l|}{\textbf{Lion}}
&
\parbox[c]{13cm}{
The correct answer is (B). The statement "Hey just be happy that you won't be depressed anymore" is sarcastic because it implies that being happy is an easy and straightforward solution to depression, which is a complex mental health issue that requires professional help and support. The use of the word "just" also adds a dismissive tone to the statement, suggesting that the speaker doesn't take depression seriously. It's important to avoid using sarcasm when talking about mental health issues, as it can be hurtful and invalidating to those who are struggling. Instead, it's important to offer compassion and support and to encourage individuals to seek professional help if needed. \textcolor{red}{\Large\textbf{\ding{55}}}
} 
\\ \midrule

\multicolumn{1}{l|}{\textbf{Vicuna}}
&
\parbox[c]{13cm}{
(B) Hey just be happy that you won't be depressed anymore \textcolor{red}{\Large\textbf{\ding{55}}}
} 
\\ \midrule

\multicolumn{1}{l|}{\textbf{ChatGPT}}
&
\parbox[c]{13cm}{
The sarcastic statement is (A) "Hey just be happy then you won't be depressed anymore." \textcolor{green}{\Large\textbf{\ding{51}}}
} 
\\

\bottomrule
\end{tabular}
\caption{A snarks case from BIGBench Hard.}
\label{tab:case8}
\renewcommand{\arraystretch}{1}
\end{table*}

\begin{table*}[!ht]
\small
\centering
\renewcommand{\arraystretch}{1}
\begin{tabular}{ll}
\toprule
\multicolumn{2}{l}{
\parbox[c]{14.3cm}{
\textbf{Instruction:} Q: In the following sentences, explain the antecedent of the pronoun (which thing the pronoun refers to), or state that it is ambiguous. Sentence: The scientist collaborated with the artist, and he shared a story. Answer Choices: (A) The scientist shared a story (B) The artist shared a story (C) Ambiguous A: Among A through C, the answer is
}
}\\
\toprule
\multicolumn{1}{l|}{\textbf{Lion}}
&
\parbox[c]{13cm}{
In the given sentence, the antecedent of the pronoun "he" is ambiguous. "He" could refer to either the scientist or the artist, as both of them shared a story. Therefore, option (C) is the correct answer. \textcolor{green}{\Large\textbf{\ding{51}}}
} 
\\ \midrule

\multicolumn{1}{l|}{\textbf{Vicuna}}
&
\parbox[c]{13cm}{
(C) Ambiguous
\\ \\
\#\#\# Explanation:
\\ \\
The pronoun "he" is ambiguous, as it could refer to either the scientist or the artist. Without more context, it is impossible to determine which person is being referred to. Therefore, the answer is "C - Ambiguous." \textcolor{green}{\Large\textbf{\ding{51}}}
} 
\\ \midrule

\multicolumn{1}{l|}{\textbf{ChatGPT}}
&
\parbox[c]{13cm}{
(C) Ambiguous.
\\ \\
In the given sentence, the antecedent of the pronoun "he" is unclear. It could refer to either the scientist or the artist because there is no clear indication of which person shared the story. Therefore, the sentence is ambiguous. \textcolor{green}{\Large\textbf{\ding{51}}}
} 
\\

\bottomrule
\end{tabular}
\caption{A disambiguation QA case from BIGBench Hard.}
\label{tab:case9}
\renewcommand{\arraystretch}{1}
\end{table*}